\pdfoutput=1
\documentclass[11pt]{article}

\usepackage[]{acl}

\usepackage{times}
\usepackage{latexsym}

\usepackage[T1]{fontenc}
\usepackage[utf8]{inputenc}

\usepackage{microtype}

\usepackage{inconsolata}

\usepackage{CJK}
\usepackage{caption}
\usepackage{subcaption}
\usepackage{graphicx}
\usepackage{multicol}
\usepackage{multirow}
\usepackage{booktabs}
\usepackage{amsmath}
\usepackage{upgreek}
\usepackage{footmisc}
\usepackage{tablefootnote}
\usepackage{footnote}
\usepackage{tabularx}
\usepackage{diagbox}
\usepackage{fancyvrb}
\usepackage{spverbatim}
\usepackage{nicematrix}
\usepackage{colortbl}
\usepackage{authblk}
\usepackage{xcolor}
\definecolor{orangeish}{HTML}{FFCC99}
\definecolor{blueish}{HTML}{87C2FF}
\newcommand\Tstrut{\rule{0pt}{2.6ex}}         % = `top' strut
\newcommand\Bstrut{\rule[-0.9ex]{0pt}{0pt}}   % = `bottom' strut
\usepackage{paralist}
\usepackage{ifsym}

\usepackage[most]{tcolorbox}
\usepackage{float}
\usepackage{xspace}
\tcbset{
  aibox/.style={
    width=474.18663pt,
    top=10pt,
    colback=white,
    colframe=black,
    colbacktitle=black,
    enhanced,
    center,
    attach boxed title to top left={yshift=-0.1in,xshift=0.15in},
    boxed title style={boxrule=0pt,colframe=white,},
  }
}
\newtcolorbox{AIbox}[2][]{aibox,title=#2,#1}

\newcommand\modified[1]{\textcolor{black}{#1}}

\newcommand\methodname{ANAH}
\title{ANAH: Analytical Annotation of Hallucinations in Large Language Models}

\author{
    Ziwei Ji$^{1,2*}$\quad Yuzhe Gu$^{1}$\thanks{Equal contributions}\quad Wenwei Zhang$^{1\dag}$\quad Chengqi Lyu$^{1}$\quad
    Dahua Lin$^{1,3}$\quad Kai Chen$^{1 \dag}$ \\
    \small{$^1$Shanghai AI Laboratory}\quad \small{$^2$Hong Kong University of Science and Technology} \\
    \small{$^3$The Chinese University of Hong Kong} \\
    \small\texttt{zjiad@connect.ust.hk}\quad \small\texttt{\{guyuzhe,zhangwenwei,lvchengqi,chenkai\}@pjlab.org.cn} 
}
\begin{document}
\maketitle
\def\thefootnote{$\dag$}\footnotetext{Corresponding author}\def\thefootnote{\arabic{footnote}}
\begin{abstract}
Reducing the `\textit{hallucination}' problem of Large Language Models (LLMs) is crucial for their wide applications. 
A comprehensive and fine-grained measurement of the hallucination is the first key step for the governance of this issue but is under-explored in the community.
Thus, we present \methodname{}, a bilingual dataset that offers \textbf{AN}alytical \textbf{A}nnotation of \textbf{H}allucinations in LLMs within Generative Question Answering.
Each answer sentence in our dataset undergoes rigorous annotation, involving the retrieval of a reference fragment, the judgment of the hallucination type, and the correction of hallucinated content. 
\methodname{} consists of $\sim$12k sentence-level annotations for $\sim$4.3k LLM responses covering over 700 topics, constructed by a human-in-the-loop pipeline.
Thanks to the fine granularity of the hallucination annotations, we can quantitatively confirm that the hallucinations of LLMs progressively accumulate in the answer and use \methodname{} to train and evaluate hallucination annotators. We conduct extensive experiments on studying generative and discriminative annotators and show that, although current open-source LLMs have difficulties in fine-grained hallucination annotation, the generative annotator trained with \methodname{} can surpass all open-source LLMs and GPT-3.5, obtain performance competitive with GPT-4, and exhibits better generalization ability on unseen questions.\footnote{Please find the dataset, code, and model at \url{https://github.com/open-compass/ANAH}.}
\end{abstract}

% Hallucination or conflation is known as the most common problem in Large Language Models (LLMs), undermining greatly the trustworthiness of the latter. 
% Despite the collective effort of the research community to mitigate this problem in LLMs, it is still not very clear how hallucinations occur in a sentence.
% We propose an analytical and fine-grained annotation of the phenomenon in a data set called ANNAH (ANAlytical Annotation of Hallucinations) via a generative question-answering task. For any given LLM-generated answer (A), we retrieve the reference answer (R) separately by a machine-assisted human process, and then we compute the entailment between the generated answer and the reference answer. 

\section{Introduction}
\label{sec: intro}

\begin{figure}[!ht]
 \centering
 \includegraphics[width=1\linewidth]{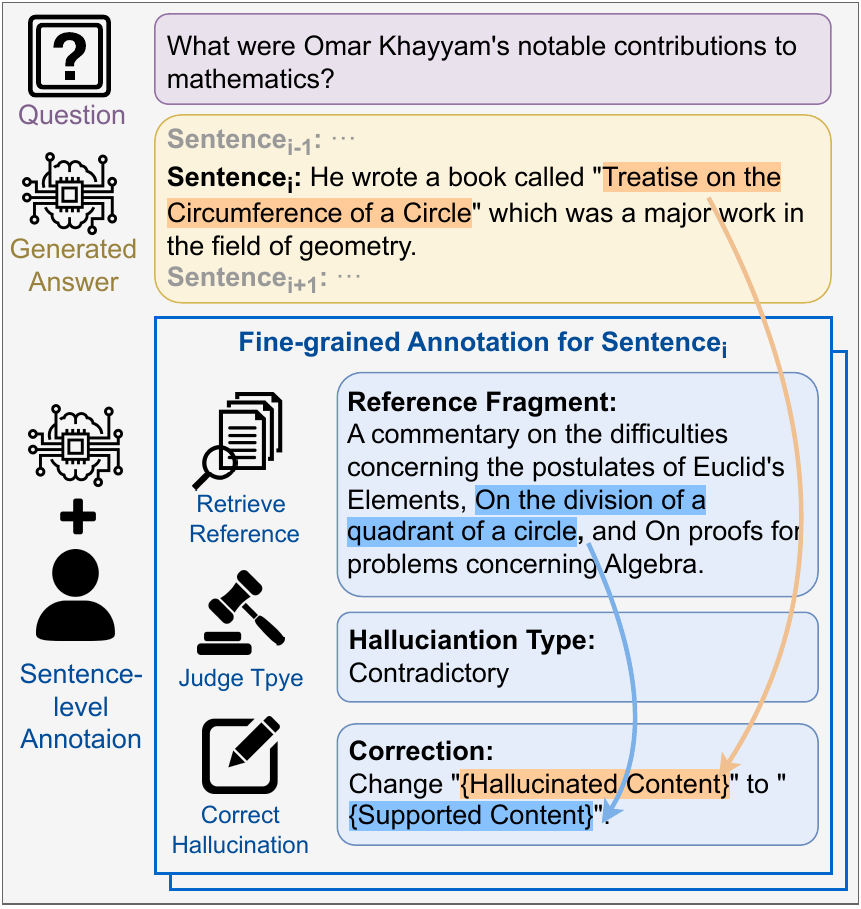}
 \vspace{-12pt}
  \caption{An example of \methodname{} for sentence-level hallucination annotation. Each sentence in a generated answer is annotated in fine-grained with Reference Fragment, Hallucination Type, and Correction. 
  The hallucinated and supported content are highlighted in \colorbox{orangeish}{orange} and \colorbox{blueish}{blue}, respectively.}
  \label{fig:annotation_example}
  \vspace{-12pt}
\end{figure}

% LLM 进展迅速，但是普遍存在幻觉，影响了真实性和应用
% With scaling up model size and data size,
Large Language Models (LLMs) have achieved significant performance improvements across a diverse array of Natural Language Processing tasks~\cite{petroni2021kilt, Kamalloo2023Evaluating, sun2023head, chen2023t, chen2024agent}. 
However, LLMs still face a worrisome problem that significantly hinders their real-world applications, \textit{hallucination}, in which they produce plausible-sounding but unfaithful or nonsensical information~\cite{ji2022survey, bang2023multitask} when answering the user questions, especially those require intensive knowledge.
% This problem curtails the performance of models in NLP tasks like Generative Question Answering (GQA) and raises safety concerns for realistic applications~\cite{su-etal-2023-context}.
% One of the representative tasks is GQA, which not only provides relevant information to answer queries, but is also widely used to evaluate the capabilities of language models.
% speciality for LLMs
Given the fluency and convincing nature of the responses produced by LLMs, the detection of their hallucinations becomes increasingly difficult~\cite{Adlakha2023Evaluating, Ren2023Investigating, Pezeshkpour2023Measuring}.
Such a challenge impedes the deep analysis and reduction of LLM hallucination and leads to extensive dissemination of misleading information as the user base widens and real-world applications proliferate~\cite{Mallen2023When}.
% 分辨不出幻觉的严重后果 传播 误导
% it becomes imperative to understand and address these issues.
% As we make AI more capable, we may get more dangerous and more likely that we lose control.

% 要去解决这个问题，首先要完整检测出LLM回复的幻觉，以进行进一步的分析或幻觉消除的工作。
% 因此如何评估幻觉检测器的能力也非常重要。需要这样的benchmark
% 然而已有工作 ABCD，已有detection的局限：（仅限于英文，或者仅侧重于检查幻觉是否存在；有些数据量虽然大，但是并不来自真实自然的 LLM 回复。都会阻碍分析
% \ziwei{high-level， 第一页结束.  why fine-tune?}
There have been extensive efforts on effectively detecting and evaluating hallucination~\cite{durmus2020feqa, Muendler2023Self, Du2023Quantifying}. 
 % Several benchmarks have been proposed to facilitate the analysis of models' ability to detect and recognize hallucinations.
 However, most benchmarks were proposed before the advent of LLM and targeted specific English tasks~\cite{dziri2021evaluating,rohrbach2018object}, which are not challenging for current models.
  % except HalluQA (LLM)
% Among recent LLM benchmarks~\cite{li2023halueval,li2024dawn}, artificially guide LLMs to produce hallucinatory unnatural responses (无法反应真实世界 开源？) and utilize questions from existing datasets. 
% The smaller Chinese Benchmark, HalluQA~\cite{cheng2023evaluating} does not focus on annotation. 短回复
Recent benchmarks~\cite{li2023halueval,li2024dawn} for LLMs only categorize whether the entire response contains hallucinations without explanation and reference. This coarse-grained nature makes it difficult to trace the exact trigger of hallucinations and obstructs further mitigation of them. 
% 进一步的分析和溯因 不能很直接帮助到消除
% Thus, they fall short of providing comprehensive insights that could facilitate further advancements in AI systems.
% Thus, such lack of detailed analysis, unnatural responses, and signal language inhibit further development in hallucination detection research.
% These limitations hinder the analysis of hallucination detection.
% Therefore, there is a lack of annotation-oriented benchmarks designed for hallucinations in LLMs that cover Chinese data and offer fine-grained annotation and reference.
 % based on their own parameterized knowledge
% thereby solving the problem holistically. Moreover, they provide a comprehensive perspective that aids in effectively training AI systems for better results.

% \ziwei{核心是dataset 不是benchmark，
% 特点 与缺点一一解决局限性}
 % 我们贡献了一个 benchmark，中英双语，并且提供了细粒度的幻觉标记，到每句话。同时都采集来自真实的模型回复，一个基于 RAG，一个基于模型的直接回答。
 % 第三段稍微展开讲清楚细粒度这个事情，引用图1，讲清楚“细”体现在哪几点
% To fill this research gap, 
Therefore, we establish a novel large-scale Chinese-English benchmark, named \textbf{\methodname{}}\footnote{\textbf{\methodname{}} is short for \textbf{AN}alytical \textbf{A}nnotation of \textbf{H}allucinations.}, that assesses the LLMs' ability to annotate the LLM hallucinations sentence-by-sentence, in the scenario of knowledge-based generative question answering.
Rather than solely result-oriented, for each answer to a question, our approach prompts the model to annotate hallucination for \textbf{each sentence}, including retrieving \textbf{reference fragment} for the sentence, judging the \textbf{hallucination type} (No/Contradictory/Unverifiable Hallucinations, and No Fact), and \textbf{correcting} the sentence based on the reference fragment if hallucination exists (Fig.~\ref{fig:annotation_example}).

% 第四段讲数据集的丰富性，需要引用图2，可以稍微补充下每一个操作的目的/价值，比如为啥是两种回复之类的。
To facilitate the scale-up of datasets, we ensure the comprehensiveness and diversity of \methodname{} across various topics, questions, and answers.
As shown in Fig.~\ref{fig: pipeline}, first, we curate topics in both English and Chinese, encompassing a broad domain range including things, places, people, and historical events (Fig.~\ref{fig:topic_distribution}).
Second, we craft around three related questions for each topic to ensure originality and avoid contamination.
% To ensure originality and avoid contamination~\cite{li2023task}, we create new questions rather than reusing existing datasets.
Third, for each question, we construct a high-quality and a low-quality response with and without reference in generation, respectively, enabling a comparative analysis of hallucination distributions across different response scenarios.
The final and pivotal stage is fine-grained hallucination annotation, as exemplified in Fig.~\ref{fig:annotation_example}.
Eventually, we form $\sim$12k hallucination annotations of $\sim$4.3k answers to $\sim$2.2k questions spanning a broad domain range, which is challenging for hallucination detection.

% Unlike HaluEval, we allow LLMs to generate naturally without intervening in the answer generation.

% Our benchmark evaluates the accuracy and coverage of knowledge within the model and the accuracy of making responses within the pre-trained knowledge demonstrates the model's ability to understand, knowledge calling, reasoning, and surface realization of information seen in various domains.
% Our benchmark evaluates the coverage of the model's internal knowledge and focuses on the hallucination level of the model in making responses within the pre-trained knowledge, reflecting their understanding, knowledge invocation, reasoning, and expressiveness across varied domains. 

% \ziwei{不要解释 不用太细 
% 为了解决上段的问题 motivation（重点）干了有什么好处
% （1）为了丰富性 
% （2）为了相关性 事实性问题
% （3）评估对比两种类型
% （4）为了以后需求 未来fine-tuned PPO
% }
% 数据构造的 pipeline
\if 0
As shown in Fig.~\ref{fig: pipeline}, the establishment of this dataset involves four key phases:
% (1) Topic Selection and Reference Retrieval:
The first phase involves the careful selection of topics and reference retrieval. To facilitate diversity and broaden the spectrum of information, we select celebrities, events, locations, and things across various domains, as in Fig.~\ref{fig:topic_distribution}. 
% (2) Question Generation and Selection: 
The second stage involves generating questions based on reference documents and selecting them on authenticity, answerability, difficulty, and variety. To ensure originality and avoid contamination~\cite{li2023task}, we create new questions rather than reusing existing datasets.
This meticulous approach guarantees a comprehensive and thought-provoking collection.
% For the factual questions, we generate multiple questions based on the reference documents, filter out similar questions, and select the final questions based on authenticity, answerability, difficulty, and variety.
% (3) Answer Generation:
The following phase is answer generation. 
To gain a comparative analysis of different hallucination distributions across different scenarios, we utilize various models to create responses both with and without references respectively. 
% RAG更准些 两种不同类型分布 也是diversity
% This methodology allows us to gain deeper insights into the functionality and effectiveness of each model.
% To compare different hallucination levels under different scenarios, we generate answers using different models with and without references, respectively. 
% (4) Fine-grained Hallucination Annotation:
The final and key stage carries out fine-grained hallucination annotations. 
This detailed annotation process enables us to not only provide a comprehensive perspective but to also further fine-tune Predictive Processing Optimization (PPO). 
Through this step, we noticed that hallucinations progressively accumulate in the responses.
% To provide a comprehensive perspective and further fine-tuned PPO, we annotate answers fine-grainedly.
% Through this process, we find hallucinations accumulate progressively in the responses.

\else

%  \ziwei{dataset好处 价值
%  我们可以先评估 后训练
% 重点结论 而非做了什么 观察到了什么
% 效果不好 所以train
% 泛化性实验 指明扩大data的方向 topic/question
% }
% 基于这个数据集训了 annotator，有哪些观察
Thanks to the completeness and fine-granularity of \methodname{}, the statistical results of the hallucination annotations quantitatively confirm that hallucinations progressively accumulate in the LLM responses.
Furthermore, \methodname{} can be used to train and evaluate hallucination annotators. We first discovered that only GPT-4 could do this task well.
Thus, we further investigate training generative and discriminative hallucination annotators using \methodname{} and observe the advantages of generative annotators over discriminative annotators in handling the imbalance issue of hallucination types.
% The discriminative annotator provides sentence-level signals of hallucinations and thus has the potential to be applied in fine-grained Reinforcement Learning from human feedback (RLHF)~\cite{wu2023fine} for mitigating hallucinations further.
Remarkably, our generative annotators achieve an accuracy of 81.01\%, surpassing open-source models and rivaling GPT-4 (86.97\%) in performance with a smaller size and lower source cost.
We also observe that the hallucination annotators consistently exhibit better generalization regarding the number of questions than the breadth of topics, thereby guiding us toward prioritizing data scaling to cover a broader array of topics in future research.

\section{Dataset Construction}

\begin{figure}[t]
 \centering
 \includegraphics[width=1\linewidth]{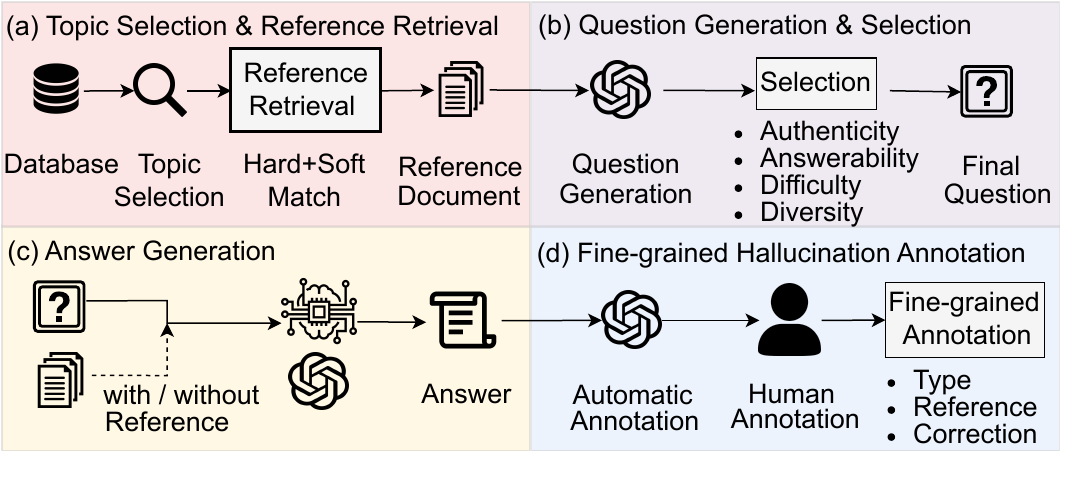}
 \vspace{-12pt}
  \caption{The overview of dataset establishment, comprising (a) Topic Selection and Reference Retrieval, (b) Question Generation and Selection, (c) Answer Generation, and (d) Fine-grained Hallucination Annotation.}
  \label{fig: pipeline}
  \vspace{-12pt}
\end{figure}

% \begin{figure}[!ht]
%      \centering
%      \begin{subfigure}[b]{1\linewidth}
%          \centering
%          \includegraphics[width=\linewidth]{figures/topic_distribution_a.png}
%          \caption{The topic distribution in categories (inner) and domains (outer).}
%          \label{fig:y equals x}
%      \end{subfigure}
%      \\
%      \begin{subfigure}[b]{1\linewidth}
%          \centering
%          \includegraphics[width=\linewidth]{figures/word_cloud.png}
%          \caption{Word Cloud.}
%          \label{fig:three sin x}
%      \end{subfigure}
% \end{figure}

% \ziwei{method每个操作都要有motivation 好处 避免和intro重复}
\methodname{}'s establishment contains four stages (Fig.~\ref{fig: pipeline}): (1) selecting a broad range of topics to ensure comprehensiveness (\S~\ref{sec:topic_selection}), (2) constructing related questions whose responses can be fully supported by reference (\S~\ref{sec:question_selection}), (3) generating answers from LLMs under different models and scenarios (\S~\ref{sec:answer_generation}), and (4) fine-grained hallucination annotation for further analysis and mitigation (\S~\ref{sec:finegrained_annotation}).

\begin{figure}[t] %[!ht]
 \centering
 \includegraphics[width=0.95\linewidth]{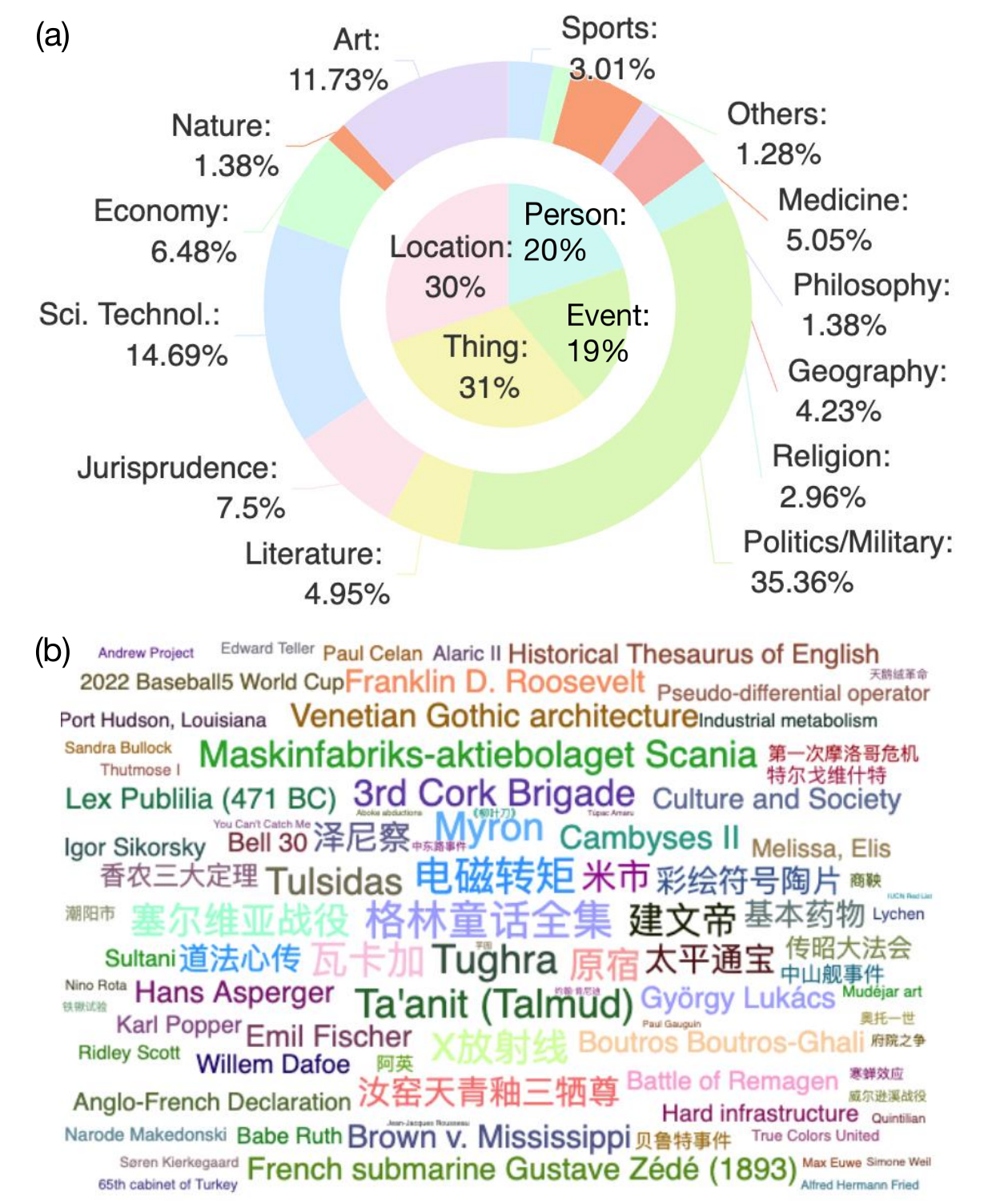}
 \vspace{-3pt}
  \caption{The topic distribution by chart of (a) categories (inner) and domains (outer), and (b) word cloud.}
  \label{fig:topic_distribution}
 \vspace{-12pt}
\end{figure}

\subsection{Topic Selection and Reference Retrieval}\label{sec:topic_selection}
% topic
The initial stage involves the selection of topics and corresponding references from knowledge-intensive datasets.
To ensure diversified and wide-ranging information, our topic choices are categorized into celebrities, events, locations, and things. We also encompass various domains, including but not limited to Politics and Military, Art, Science and Technology, Religion, \textit{etc}. ( Fig.~\ref{fig:topic_distribution}).
\modified{Topics are meticulously chosen based on the frequency of their occurrence via Google Ngram Viewer~\footnote{\url{https://books.google.com/ngrams/}} since topics that more frequently occur and are of public interest should be more important for real-world applications of LLMs.
We also collect topics from publicly available summaries like historical timelines and the ranking of influential persons~\footnote{For example, \url{https://en.wikipedia.org/wiki/Timeline_of_Chinese_history} and \url{https://pantheon.world/explore/rankings}.}.
}

% retrieval
After selecting the topics, their corresponding reference documents are retrieved from pre-training databases~\cite{conghui2022opendatalab}, including Wikipedia\footnote{\url{https://www.wikipedia.org/}}, Baidu Baike\footnote{\url{https://baike.baidu.com/}}, Encyclopedia Britannica\footnote{\url{https://www.britannica.com/}}. 
% Encyclopedia of China~\footnote{\url{https://www.zgbk.com}},
We select the datasets that have been widely used in the pre-training stage of LLMs~\cite{touvron2023llama} so that we can make sure that the model saw the truth, which is important for further analysis and mitigation of hallucinations.

During the reference retrieval process, the discrepancies in nomenclature across different sources and the potential of a single name having multiple meanings present challenges. 
To address these challenges, we adopt a strategy that progresses from hard to soft matching.
First, we perform exact matching (i.e., hard matching) of the entries.
Then, we sort the candidate entries according to the sentence semantic similarity and further judge them with InternLM~\cite{2023internlm} to select the correct ones~\footnote{Please refer to Fig.~\ref{fig: reference retrieval prompt} for the prompt.}.
Finally, manual filtering is performed to iron out the problem of renaming.
Overall, this phase establishes a robust foundation for the ensuing steps of benchmark construction.

\subsection{Question Generation and Selection}\label{sec:question_selection}
The second stage involves the generation and selection of several questions based on the provided reference documents about a particular topic. 
% Generation
% free from contamination
To increase the possibility that the data is unseen and untainted, we create new questions rather than repurposing existing datasets.
The questions are framed in a manner so that they can be fully answered exclusively grounded on the provided reference documents, avoiding being overly subjective or open-ended.
To ensure diversity and comprehension across questions, they are designed to cover different types, such as `what', `when', `where', `why', etc, and perspectives such as descriptions, explanations, reasons, etc., encapsulating all facets of the information.
The questions also traverse diverse levels of knowledge, ranging from basic, generic knowledge to more intricate, specialized knowledge or domain-specific expertise.
The generation prompt is shown in Fig.~\ref{fig: QG prompt}.

\begin{table*}[ht]
\centering
 \resizebox{0.95\linewidth}{!}{
\begin{tabular}{ll}
\toprule
\textbf{Topic} & \multicolumn{1}{c}{Omar Khayyam}  \\ \hline
\textbf{Reference}   & \begin{tabularx}{1\textwidth}{X}
Ghiy\=ath al-D\={\i}n Ab\=u al-Fatḥ Umar ibn Ibr\=ah\={\i}m N\={\i}s\=ab\=ur\={\i}, commonly known as Omar Khayyam, was a polymath, known for his contributions to mathematics, astronomy, philosophy, and {\color[HTML]{9400D3} $^{0}$ Persian poetry}...
{\color[HTML]{9400D3} $^{0}$ As a mathematician, he is most notable for his work on the classification and solution of cubic equations, where he provided geometric solutions by the intersection of conic}...

\#\# Mathematics  

Khayyam was famous during his life as a mathematician. {\color[HTML]{3166FF} $^{1}$ His surviving mathematical works include: A commentary on the difficulties concerning the postulates of Euclid's Elements, On the division of a quadrant of a circle, and On proofs for problems concerning Algebra}...

\#\#\# Real number concept

Khayyam discusses the relationship between the concept of ratio and the concept of number and explicitly raises various theoretical difficulties. {\color[HTML]{006400} $^{2}$ In particular, he contributes to the theoretical study of the concept of irrational number. }...\end{tabularx}   \\\hline
\textbf{Question}   & \begin{tabularx}{1\textwidth}{X} What were Omar Khayyam's notable contributions to mathematics? \end{tabularx} \Tstrut\Bstrut  \\  \hline
\textbf{Answer}  &  \begin{tabularx}{1\textwidth}{X}
{\color[HTML]{9400D3} $^{0}$ 
Omar Khayyam was a mathematician and poet who is best known for his contributions to algebra and geometry.}
{\color[HTML]{3166FF} $^{1}$ He wrote a book called "Treatise on the Circumference of a Circle" which was a major work in the field of geometry.}
{\color[HTML]{006400} $^{2}$ He is also known for his work on the theory of numbers and for his contributions to the development of the decimal system.}
{\color[HTML]{808080}$^{3}$ I hope my reply is helpful.}
\end{tabularx} \Bstrut \\ \hline

\textbf{\begin{tabular}[c]{@{}l@{}}Sent 0\\ Annotation\end{tabular}} &  \begin{tabularx}{1\textwidth}{X}  \textbf{<Reference>} Ghiy\=ath al-D\={\i}n Ab\=u al-Fatḥ Umar ibn Ibr\=ah\={\i}m N\={\i}s\=ab\=ur\={\i}, commonly known as Omar Khayyam, was a polymath, known for his contributions to mathematics, astronomy, philosophy, and  Persian poetry.<SEP>
As a mathematician, he is most notable for his work on the classification and solution of cubic equations, where he provided geometric solutions by the intersection of conic.

\textbf{<Halluciantion>} None   
 \end{tabularx}  \\ \hline

\textbf{\begin{tabular}[c]{@{}l@{}}Sent 1\\ Annotation\end{tabular}} &  \begin{tabularx}{1\textwidth}{X}  \textbf{<Reference>} A commentary on the difficulties concerning the postulates of Euclid's Elements, On the division of a quadrant of a circle, and On proofs for problems concerning Algebra. 

\textbf{<Halluciantion>} Contradictory    

\textbf{<Correction>}  "Treatise on the Circumference of a Circle" to "On the division of a quadrant of a circle".
 \end{tabularx} \Tstrut\Bstrut \\  \hline
\textbf{\begin{tabular}[c]{@{}l@{}}Sent 2\\ Annotation\end{tabular}} & \begin{tabularx}{1\textwidth}{X}  
\textbf{<Reference>} In particular, he contributes to the theoretical study of the concept of irrational number. 

\textbf{<Halluciantion>} Unverifiable    

\textbf{<Correction>}  "and for his contributions to the development of the decimal system." to "".
 \end{tabularx} \Tstrut\Bstrut \\\hline
 \textbf{\begin{tabular}[c]{@{}l@{}}Sent 3\\ Annotation\end{tabular}} &  \begin{tabularx}{1\textwidth}{X}  \textbf{<No Fact>} \end{tabularx}  \\ \bottomrule   
\end{tabular}
}
\vspace{-3pt}
  \caption{Examples of fine-grained hallucination annotation for each sentence in an answer. Related fragments for each sentence in reference are marked in the same colors ({\color[HTML]{9400D3}purple}, {\color[HTML]{3166FF}blue}, {\color[HTML]{006400}green}, and {\color[HTML]{808080} grey} for sentence 0, 1, 2, and 3, respectively). }
  \label{tab: example}
  \vspace{-12pt}
\end{table*}

% Selection
To assure the uniqueness of each question and avoid duplication, we leverage CoSENT~\footnote{\url{https://huggingface.co/shibing624/text2vec-base-chinese}} for Chinese and MiniLM~\footnote{\url{https://huggingface.co/sentence-transformers/all-MiniLM-L6-v2}} for English, respectively, to calculate similarities among questions and sift out overly similar ones\footnote{Pleaser refer to Appendix~\ref{appendix sec: prompt} for details.}.
We then employ GPT-3.5~\cite{openai_2023} to assess their answerability, 
\textit{i.e.}, whether the given questions can be answered based solely on the provided reference documents.
This ensures that the questions are fact-based, objective, and possess a definitive answer, thus increasing the reliability and consistency of the evaluation process.
The prompt details are in Fig.~\ref{fig: answerability prompt}.

Finally, we utilize GPT-4~\cite{openai_2023} to select the top three questions from the \modified{above candidate questions}, considering the following characteristics:
\begin{compactenum}
\item High authenticity: The questions should be free from any intentionally misleading, ambiguous, or false information.
\item High answerability: The questions exhibiting excessive subjectivity, controversy, or predictive nature should be excluded.
\item Difficulty: A certain level of difficulty should be guaranteed.
\item High diversity: Enhancement of overall diversity in terms of type, complexity, depth of knowledge, \textit{etc}. Similar questions should be discarded.
\end{compactenum}
The question selection prompt is in Fig.~\ref{fig: best q prompt}.
This meticulous process of question generation and selection not only ensures the quality of the benchmark but also elevates its value in testing the model hallucinations.

\subsection{Answer Generation}\label{sec:answer_generation}
The third stage involves generating answers for each question with different LLMs. 
In this case, we use GPT-3.5 with a reference document to construct a high-quality answer and an early version of InternLM-7B without reference to generate a low-quality answer, respectively.
Such a design allows to evaluation of the LLMs’ hallucination annotation capability under different scenarios comprehensively.
% Multiple-scenario generation allows us to explore the different hallucination distributions 
% and derive valuable insights into the knowledge acquisition, contextual understanding, and reasoning capabilities of these LLMs.
% deepen our comparative analysis to 
Please refer to Fig.~\ref{fig: answer prompt} for details of answer generation with reference.

\subsection{Fine-grained Hallucination Annotation}
\label{sec:finegrained_annotation}
% 应该先 ref table 1 讲怎么进行的细粒度标注流程（需要先从原本的文档里通过 embedding 模型找到对应的 reference fragment，然后基于 reference fragment 判断 hallucination type，再correct）。讲完这个流程以后，可以说这个标注太耗时了，所以需要引入 GPT-4 辅助标注，并对reference fragment 的retrieval进行优化

The final stage involves fine-grained hallucination annotation for the answers to each question generated in the previous stages.
As shown in Tab.~\ref{tab: example}, we provide the annotators with documents on a specific topic and a related question.
For each answer sentence, the complete annotation includes finding the exactly related reference fragments, assessing the hallucination type, and correcting the hallucinations accordingly. 
To reduce the extensive time and human labor\footnote{typically 20 minutes per answer per annotator.} and keep accuracy, we adopt GPT-4~\cite{openai_2023} for preliminary annotation, followed by the verification and refinement of human annotators.

% window
% retrieval 的优化目标有两个，一个是尽可能提高覆盖率，即需要准确找到覆盖reference的那段文本，同时让这段覆盖文本的长度尽量短，以提升标注效率并交给使用gpt4时导致的文本context过长。所以用了若干模型提升精度，并且empirically find 4k还是8k就足够了
% Considering the constraints of maximum input length and cost of using GPT-4,
Specifically, we first apply existing retrieval methods to determine a document window for each answer sentence that accurately encapsulates related information.
We empirically choose BM25~\cite{robertson2009probabilistic} for both language, and further apply two CoSENT mdoels\footnote{\url{https://huggingface.co/shibing624/text2vec-base-chinese} and \url{https://huggingface.co/shibing624/text2vec-bge-base-chinese}} for Chinese, and MiniLM\footnote{\url{https://huggingface.co/sentence-transformers/all-MiniLM-L6-v2}} for English, to rank reference fragments. 
The ensemble of multiple embedding models significantly improves retrieval accuracy, which serves as a foundation for accurate hallucination-type classification and hallucination correction and reduces the cost of human annotators to correct the reference fragment.
Furthermore, to optimize resource utilization of GPT-4 without compromising the annotation accuracy, we empirically determine the context length of reference fragments to be 540 tokens for Chinese and 400 tokens for English. For the remaining unverifiable sentences due to the failure of retrieval, we extend the window length by sixfold for secondary annotation and finally fix the remaining cases after secondary annotation by human annotation. 

% annotation
Based on the document window for each answer sentence, GPT-4 is prompted to identify reference fragments and assess whether hallucinations exist. If the sentence contains factual information and aligns with the reference, its type is `No Hallucination'. Annotators should also pinpoint the specific reference fragments from the original documents. 
If the sentence contradicts the reference, its type is `Contradictory Hallucination'. The specific reference fragments and a suggestion on correcting the response are required.
If the sentence lacks supporting evidence and cannot be verified, its type is `Unverifiable Hallucination' and a revision suggestion is required.
If the sentence does not contain any factual information for evaluation, it falls under the category of `No Fact' without further annotation.
See detailed GPT-4 prompts in Fig.~\ref{fig: ann prompt0}.
After preliminary annotation, human annotation is conducted following a similar workflow.

\begin{table}[t]
\centering
\resizebox{1\linewidth}{!}{
\begin{tabular}{ccccc}
\toprule
\multicolumn{1}{c}{\textbf{Language}} & \multicolumn{1}{c}{\textbf{\begin{tabular}[c]{@{}c@{}}\# Topic\end{tabular}}} & \multicolumn{1}{c}{\textbf{\begin{tabular}[c]{@{}c@{}}\# Ans\end{tabular}}} & \multicolumn{1}{c}{\textbf{\begin{tabular}[c]{@{}c@{}}\# Sent\end{tabular}}}  & \multicolumn{1}{c}{\textbf{\begin{tabular}[c]{@{}c@{}}\# Token (w/,w/o Ref)\end{tabular}}} \\ \hline
English &  476  &  2,626 & 6,606 & 4.1M / 642K \\
Chinese &  324 &  1,772  & 5,582 & 2.8M / 683K \\ \bottomrule
\end{tabular}
}
\vspace{-3pt}
\caption{Number of topics, annotated answers, annotated sentences, and tokens (with and without reference documents) for each language of \methodname{}.}
\label{tab:number_data}
\vspace{-12pt}
\end{table}

\subsection{Dataset Statistics}
Eventually, our dataset covers both English and Chinese and comprises over 700 topics, $\sim$4.3k annotated answers, $\sim$12k annotated sentences, and $\sim$7M tokens with reference documents (Tab.~\ref{tab:number_data}). The topics also cover celebrities, events, locations, and things, from an array of domains, such as military/politics, health/medicine, and sports, as depicted in Fig.~\ref{fig:topic_distribution}.
The statistics underscore the comprehensiveness and extensive scale of our dataset.

We also verify the quality of GPT-4 generated annotations by analyzing their consistency with human annotations (the higher, the better). As shown in Tab.~\ref{tab: correlation}, the average consistency is 86.97\% for hallucination type, 85.37\% for reference, and 78.98\% for correction. GPT-4 tends to erroneously annotate sentences as `No Fact' when sentences contain referential ambiguity or summary discussion, while the type of `No Fact` only accounts for $\sim$2\% of annotated sentences. 
We provide inconsistent examples in \S \ref{appendix sec: Case}.
% The high consistency shows the high quality of GPT-4 annotations.

\begin{table}[!t]
\centering
\resizebox{0.9\linewidth}{!}{
\begin{tabular}{llllll}
\toprule
\multicolumn{4}{c}{\textbf{Hallucination Type}}  & \multirow{2}{*}{\textbf{Ref}} & \multirow{2}{*}{\textbf{Corr.}} \\ \cline{1-4}
\multicolumn{1}{c}{\textbf{None}} & \multicolumn{1}{c}{\textbf{Cont.}} & \multicolumn{1}{c}{\textbf{Unver.}} & \multicolumn{1}{c}{\textbf{N.F.}} & & \\ \hline
90.19  &  83.70 &  75.69  &  28.67  &  85.37 &  78.98 \\ \bottomrule    
\end{tabular}
}
% 86.97   &  % accuracy 
\vspace{-3pt}
\caption{Consistency between GPT-4 and human Annotations 
, where `Cont.', `Unver.', `N.F.', `Ref.', and `Corr.' are abbreviations of Contradictory, Unverifiable, No Fact, Reference, and Correction, respectively.}
\label{tab: correlation}
\vspace{-12pt}
\end{table}

Tab.~\ref{tab:performance} presents the proportions of hallucination type for answers generated by GPT-3.5 with reference and InternLM without reference. The hallucination proportions for answers generated with reference are much higher than those without. Such an observation which is consistent with recent research interests in retrieval augmented generation (RAG)~\cite{lewis2020retrieval}.

\noindent\textbf{Accumulation Effect}
Thanks to the fine granularity of \methodname{}, we can quantitatively analyze the accumulation or snowball effect of hallucinations~\cite{Snowball2023}.
The probability of hallucinations occurring in the current sentence when the previous sentences contain hallucinations, $P(H_t|H_{[0:t-1]})$, is defined as 
\begin{equation}\label{equ: snowball}
\setlength{\abovedisplayskip}{4pt}%shrink space
\vspace{-3pt}
\begin{split}
P(H_t|H_{[0:t-1]}) = \frac{P(H_t, H_{[0:t-1]})}{P(H_{[0:t-1]})}, \\
\textrm{where} \quad H_{[0:t-1]} = \exists t' \in [0:t-1]:H_{t'}.
\end{split}
\end{equation}
$H_t$ is a Boolean indicator that returns true if the current sentence is hallucinated.
The hallucination probability is \textbf{58.51\%} for English and \textbf{52.54\%} for Chinese, while the hallucination probability when the previous sentences don't contain, $P(H_t|\sim H_{[0:t-1]})$, is \textbf{14.61\%} for English and \textbf{17.2\%} for Chinese.
$P(H_t|H_{[0:t-1]})$ is significantly higher than $P(H_t|\sim H_{[0:t-1]})$ indicates that the probability of hallucinations increases when the previous sentences contain hallucinations compared to when there are not, which quantitatively confirms the accumulation effect of hallucinations.
% This suggests that hallucinations tend to accumulate progressively over consecutive sentences.

\begin{table}[!t]
\centering
\resizebox{0.8\linewidth}{!}{
\begin{tabular}{llccccc}
\toprule
\multicolumn{1}{c}{\textbf{Lang}} & \multicolumn{1}{c}{\textbf{}} & \multicolumn{1}{c}{\textbf{None}} & \multicolumn{1}{c}{\textbf{Cont.}} & \multicolumn{1}{c}{\textbf{Unver.}}  & \multicolumn{1}{c}{\textbf{N.F.}} \\ \hline
\multirow{2}{*}{EN} & w/ Ref &  89.94  &  3.35  & 5.48 &  1.23 \\
                    & w/o Ref &  41.31  & 24.07  & 32.94 & 1.68 \\  \hline
\multirow{2}{*}{ZH} & w/ Ref & 74.86 & 8.04 & 16.05 & 1.05 \\
                    & w/o Ref & 31.82 & 28.07 & 35.86 & 4.25 \\ \bottomrule
\end{tabular}
}
\vspace{-3pt}
\caption{Proportion of each annotation type for answers generated with and without reference in English and Chinese.}
\label{tab:performance}
\vspace{-12pt}
\end{table}
\section{Hallucination Annotator}
\label{sec: annotator}
Taking advantage of the rich fine-grained annotations in \methodname{}, we explore training and evaluating both generative and discriminative annotators. 
The generative annotator generates textual annotations including reference fragments, hallucination type, and correction; 
while the discriminative annotator only focuses on discriminating hallucination type.

\subsection{Generative Annotator}
\label{subsec: generative annotator}
We adopt the same pipeline and prompts as the preliminary annotation of GPT-4 for the generative annotator.
We first comprehensively analyze the current open-source and close-source LLMs' ability to generate fine-grained hallucination annotation using \methodname{}.
Specifically, consistency with humans is assessed through the examination of an array of multilingual LLMs including Llama2~\cite{touvron2023llama}, InternLM2, Qwen~\cite{qwen}, Baichuan2~\cite{baichuan2023baichuan2} in different sizes, GPT-3.5, and GPT-4. 

In addition, we explore training hallucination annotators using InternLM on our dataset.
% In the early stages of benchmark establishment, we utilize GPT-4 to annotate the answers. % for annotation and replace GPT-4 to save money and improve performance. 
The fine-grained annotation involves constructing multiple sentence annotations from each answer. When constructing the training data, each sentence from an answer forms a sample.
% we can use single or multiple turns. 
% In single-turn, each sentence in an answer forms a sample, while in multiple-turns, an answer forms a sample and each sentence is a turn.

\noindent\textbf{Data Augmentation}
We perform a multi-task setting where besides fine-grained hallucination annotation, we incorporate other tasks including question generation, question selection, answer generation from intermediate products of \methodname{}, and dialogue generation from ShareGPT~\cite{ShareGPT} and Dolly~\cite{DatabricksBlog2023DollyV2}.
In addition, we apply prompt augmentation by the design of multiple prompts with varying instruction descriptions, relative locations of reference and question, etc. 
% Please refer to Fig.~\ref{fig: ann prompt0} to~\ref{fig: ann prompt4} in \S~\ref{appendix sec: prompt} for the prompts. 
Please refer to \S~\ref{appendix sec: annotation prompt} for details.

\subsection{Discriminative Annotator}
\label{subsec: discriminative annotator}
% 不同用法的打分器 
% Q+A
% Q+A+goldenA
% Q+A+D
% The discriminative annotator, as a proxy for human feedback, can discriminate whether the output is desirable and drive LLMs' weight updates in fine-grained RLHF~\cite{wu2023fine}.
Recent works~\cite{wu2023fine, lightman2023let, uesato2022solving} explore process-supervised reword models to provide fine-grained signals in RLHF, which are also useful in hallucination mitigation process such as RLHF~\cite{wu2023fine}.
% at a step-level. 
% Such a method has been proved efficient in the area of mathematical reasoning~\cite{wang2023math}, inspiring us to apply it to sentence-level hallucination detection.
Thus, we also explore training a sentence-level process-supervised discriminative annotator using InternLM, based on \methodname{}, which has the potential to be applied for fine-grained RLHF.

% Process-supervised discriminative annotators can provide denser and more accurate scoring signals than traditional outcome-supervised discriminative annotators.
% In other domains, many works \cite{} have demonstrated that fine-grained RLHF using process-supervised discriminative annotators can be effective in improving the model's capabilities.
Following the sentence-level information including references and hallucination type of \methodname{}, the model is trained to categorize each sentence into one of four types: No/Contradictory/Unverifiable Hallucination, and No Fact. 
To enable process supervision and reuse the learned knowledge in LLMs, we replace the last layer of the pre-trained LLM with a four-category linear layer and load the remaining parameters of pre-trained LLMs for further training the annotators.
This approach ensures that the scoring results are compatible with reward models in various aspects, including relevance and completeness~\cite{wu2023fine}. 
Additionally, the inference time of the discriminative annotator is significantly shorter than that of its generative counterparts.
% mix task 统一
% will be output at the end of each sentence.
% To simplify the scoring signaling, we categorize each sentence into three types: positive, negative, or neutral.
% The positive label indicates that the sentence is free of hallucinations and needs to be encouraged.
% The negative label implies that the sentence is \textit{Contradictory} or \textit{Unverifiable}, and thus does not align with our desired output. "Unverifiable" is assigned to the negative label since this judgment is based upon a limited reference, suggesting the potential presence of risks. 
% And the neutral labels indicate that the sentence contains \textit{No Fact} to judge.

\section{Experiments}
\label{sec: experiments}

\subsection{Implementation} 
\label{subsec: implement}
\noindent\textbf{Data Split}
\methodname{} is divided into training and testing sets.
To investigate the direction of annotator generalization and dataset scaling, we further divide the testing set equally into unseen-topic and unseen-question groups.
In the unseen-topic test set, the topics and corresponding references, questions, and answers remain unexposed during training.
In the unseen-question test set, the topics have been exposed during training, while the questions remain unexposed.

% To investigate the direction of annotator generalization and dataset scaling, we divide the training and testing sets based on topic and question, respectively. 
% To control for variables, both dividing methods have the same training subset and an equal number of test subsets.
% When divided by topic, the topics and corresponding references, questions, and answers in the testing data remain unexposed throughout the training phase. This way enables the examination of generalization applicable to previously unseen topics.
% On the other hand, when divided by question, the topics in the test data may have been exposed during the training phase, while the questions remain unexposed.

% In addition, we explore some training settings such as the tokenize method, data ratio, whether to calculate input loss, etc. 
Further details regarding the experimental implementation can be found in \S~\ref{appendix sec: generative ann implementation} for generative annotator and \S~\ref{appendix sec: discriminative ann implementation} for discriminative annotator.

\subsection{Evaluation Protocols}
% \ziwei{motivation 加上理由 展开一点}
\label{sec: auto_eval}
% The general idea is to use classic metrics for evaluation based on the constructed benchmark
% We assess the quality of fine-grained hallucination annotation in terms of three key elements.
For the hallucination type predicted by generative and discriminative annotators, we utilize \modified{\textbf{F1}} and \textbf{Accuracy} to measure the quality of predicted categorization.
As discriminative annotators can only classify hallucination types, we only evaluate reference fragments and corrections predicted by generative annotators and employ~\textbf{RougeL}~\citep{lin2004rouge} and \textbf{BertScore}~\cite{bert-score} to compare the generated text with gold-standard human reference in terms of gram, continuity, order and semantics.
Since we aspire that the reference sentence predicted by generative annotators originate from the document, we also apply \textbf{n-gram Precision} to reflect fidelity to the source information.
% We employ \textbf{Accuracy} to assess the capability of our discriminative annotator.
% calculating the proportion of correct feedback for different hallucination types.
% direction? prediction? signal?
% and does not yet address the subsequent reinforcement learning process like RLHF, 

\subsection{Overall Results}
\label{subsec: annotator res}
\begin{table}[!t]
\centering
\resizebox{1\linewidth}{!}{
\begin{tabular}{lcccccc}
\toprule
\multicolumn{1}{c}{\textbf{Model}} & \textbf{F1} $\uparrow$ & \textbf{ACC} $\uparrow$ & \textbf{R} $\uparrow$ & \textbf{BERT} $\uparrow$ & \textbf{Pre4} $\uparrow$ \\ \hline \Tstrut
GPT-3.5 & 48.01 & 47.94 & 29.4 & 78.78 & 64.25 \\ % & 14.3
GPT-4 & 87.11 & 86.97 & 86.32 & 96.21 & 86.44 \\ \hline % & 80.47
Qwen-7B & 8.46 & 4.67 & 24.28 & 77.28 & 44.89 \\ % & 14.27
Baichuan2-7B & 9.63 & 5.50 & 4.21 & 10.65 & 39.82 \\ % & 0.50
% Mistral-7B  &  \\ 
LLama2-7B & 13.76 & 8.31 & 4.37 & 19.93 & 8.26 \\ % & 4.68
% InternLM2-7B & 64.71 & 49.71 & 86.53 & 97.04 \\% & 37.18
InternLM2-7B & 12.44 & 12.34 & 9.54 & 64.19 & 55.72 \\ 
Qwen-14B & 14.94 & 8.82 & 10.53 & 55.2 & 85.65 \\ % & 1.81
Baichuan2-13B & 42.17 & 38.04 & 23.39 & 75.27 & 36.9 \\ % & 10.38
LLama2-13B & 8.55 & 4.80 & 5.15 & 20.16 & 13.65 \\ % & 5.85
InternLM2-20B & 61.49 & 63.17 & 46.36 & 84.68 & 94.93 \\ % & 29.54
Qwen-72B & 58.27 & 55.69 & 35.96 & 79.21 & 77.19 \\ % & 16.95
Llama2-70B & 18.42 & 12.53 & 7.13 & 20.95 & 43.31 \\ \hline% & 7.54
\methodname{}-7B & 78.69 & 79.92 & 58.51 & 87.27 & 94.90 \\
\methodname{}-20B & 80.49 & 81.01 & 58.82 & 88.44 & 94.86\\
\bottomrule
\end{tabular}
}
% GPT-3.5 & 47.94 & 47.07 & 14.3 & 20.17 & 29.39 & 32.36 & 64.25 & 49.67 \\
% GPT-4 & 86.97 & 86.19 & 81.41 & 79.54 & 88.02 & 84.62 & 90.02 & 82.87 

% Zero-shot 
% \quad\quad
\vspace{-3pt}
\caption{Automatic evaluation results for generative hallucination annotators based on different models, where `R', `BERT', and `Pre4' refer to `RougeL', `BERTScore', and `4-gram Precision', respectively.}
\label{tab:auto_result ann models}
\vspace{-6pt}
\end{table}

\noindent\textbf{Generative Annotator}
The results on the whole testing set in Tab.~\ref{tab:auto_result ann models} show current open-source LLMs and GPT-3.5 struggle to follow the instructions to annotate hallucination in a fine-grained manner, while GPT-4 exhibits high consistency with humans. 
Consequently, we train our hallucination annotators utilizing the train split of \methodname{}.
Remarkably, our \methodname{}-20B achieves \modified{an F1 of 80.49\%} and an accuracy of 81.01\%, surpassing open-source models and rivaling GPT-4 in performance with a smaller size and lower source cost. 
We notice our model exhibits higher Precision but lower RougeL than GPT-4, indicating fidelity to the original documents but inaccurate identification of reference fragments and correction.
\modified{Please refer to Tab.~\ref{tab: topic-specific performance} for topic-specific analysis of \methodname{}-7B in Appendix~\ref{appendix sec: results}.}
% Therefore, the breadth of the topic enhances the model's ability to generalize to previously unknown queries.

\begin{table}[!t]
\centering
\resizebox{1\linewidth}{!}{
\begin{NiceTabular}{lcccccccc}
\CodeBefore
  \cellcolor[HTML]{EFEFEF}{2-2,3-2,4-2,5-2,6-2,7-2,2-4,3-4,4-4,5-4,6-4,7-4,2-6,3-6,4-6,5-6,6-6,7-6,2-8,3-8,4-8,5-8,6-8,7-8}
\Body
\toprule
\multicolumn{1}{c}{\multirow{2}{*}{\textbf{Setting}}} & \multicolumn{2}{c}{\textbf{F1}$\uparrow$} & \multicolumn{2}{c}{\textbf{ACC}$\uparrow$} & \multicolumn{2}{c}{\textbf{RougeL}$\uparrow$} & \multicolumn{2}{c}{\textbf{Pre4}$\uparrow$} \\ \cline{2-9} 
\multicolumn{1}{c}{}   & \textbf{T}   & \textbf{Q}  & \textbf{T}   & \textbf{Q}   & \textbf{T}    & \textbf{Q}   & \textbf{T}      & \textbf{Q}     \\ \hline 
\begin{tabular}[c]{@{}l@{}}
G-7B\end{tabular} & 75.93 & 77.24 & 77.89 & 78.12 & 58.02 & 57.76 & 95.62 & 95.17 \\
G-20B & 79.82 & 81.18 & 80.21 & 81.81 & 56.01 & 61.62 & 94.97 & 94.77 \\ \hline
D-7B & 66.20 & 68.53 & 69.15 & 70.86 & - & - & - & - \\ 
D-20B & 69.74 & 73.98 & 72.10 & 75.95 & - & - & - & - \\
\bottomrule
\end{NiceTabular}
}
% \quad\quad
\vspace{-3pt}
\caption{Evaluation results for generative and discriminative annotators, noted by `G' and `D', respectively. `T' represents the unseen-topic test set, while `Q' represents the unseen-question test set.~\protect\footnotemark}
\label{tab:gen_and_dis_ann}
\vspace{-15pt}
\end{table}
\footnotetext{Due to the space limit, we put BERTScore in Tab~\ref{tab:gen_ann}.}

\noindent\textbf{Discriminative Annotator}
Tab.~\ref{tab:gen_and_dis_ann} shows the \modified{F1 and the accuracy} of the discriminative annotator is relatively lower than that of the generative annotator.
Thus, we analyze the confusion matrices of hallucination type for both annotators. 
Fig.~\ref{fig:cm_as_20} shows the discriminative annotator is more prone to misjudge into the largest category (No Hallucination), with the 2nd to 4th row of the 1st column totaling 255, exceeding 147 for generative annotator, given the data imbalance issue depicted in Tab.~\ref{tab:performance}.
This suggests the current discriminative annotators are more affected by the imbalance issue of hallucination types and require further modification for improvements, which we leave for future research.
Refer to \S~\ref{appendix sec: results} for all confusion matrices.

\noindent\textbf{Generalization Analysis}
Tab.~\ref{tab:gen_and_dis_ann} also indicates both generative and discriminative annotators perform better on the unseen-question test set than the unseen-topic test set in the hallucination-type classification task.
This suggests leveraging prior knowledge learned from the same topic in training aids in handling exposed references in testing.
% We deduce that this is because the annotator can leverage prior knowledge learned from the same topic in training for handling exposed references during testing.
This implies extending the breadth of topics has higher priority than extending questions of the same topic when scaling the data sizes of hallucination annotation in the future.
\modified{In addition, we assess the generalization of \methodname{} annotator to other LLMs (e.g. Qwen-7B, Baichuan2-7B) in Appendix~\ref{appendix sec: generalization on other LLMs}.}

\begin{figure}[!t]
 \centering
 \includegraphics[width=0.9\linewidth]{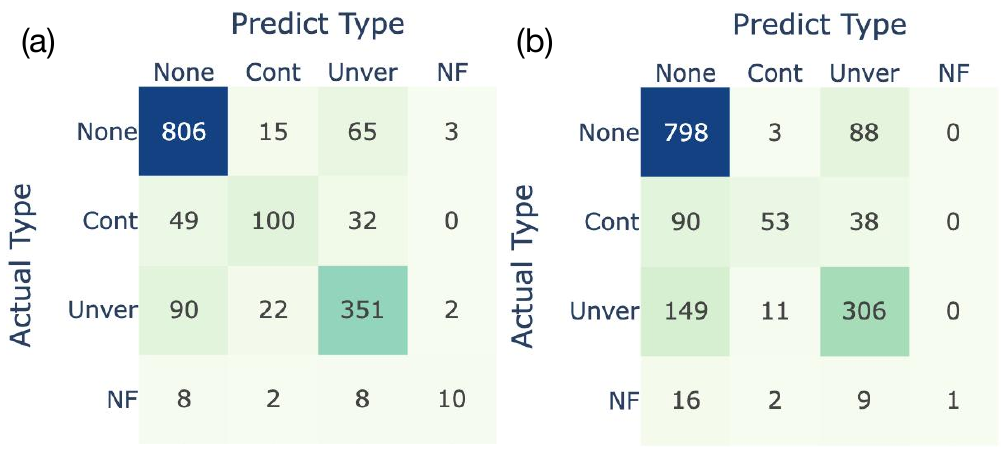}
 \vspace{-3pt}
  \caption{Hallucination Type Confusion Matrices for InternLM2-20B-based  generative annotator (a) and discriminative annotator (b).}
  \label{fig:cm_as_20}
  \vspace{-6pt}
\end{figure}

\subsection{Ablation Study}

\begin{table}[!t]
\centering
\resizebox{1\linewidth}{!}{
\begin{NiceTabular}{lcccccccc}
\CodeBefore
  \cellcolor[HTML]{EFEFEF}{2-2,3-2,4-2,5-2,6-2,7-2,2-4,3-4,4-4,5-4,6-4,7-4,2-6,3-6,4-6,5-6,6-6,7-6,2-8,3-8,4-8,5-8,6-8,7-8}
\Body
\toprule
\multicolumn{1}{c}{\multirow{2}{*}{\textbf{Setting}}} & \multicolumn{2}{c}{\textbf{F1}$\uparrow$} & \multicolumn{2}{c}{\textbf{ACC}$\uparrow$} & \multicolumn{2}{c}{\textbf{RougeL}$\uparrow$} & \multicolumn{2}{c}{\textbf{Pre4}$\uparrow$} \\ \cline{2-9} 
\multicolumn{1}{c}{}   & \textbf{T}   & \textbf{Q}  & \textbf{T}   & \textbf{Q}   & \textbf{T}    & \textbf{Q}   & \textbf{T}      & \textbf{Q}     \\ \hline 
\begin{tabular}[c]{@{}l@{}}
S.T.\end{tabular} & 75.93 & 77.24 & 77.89 & 78.12 & 58.02 & 57.76 & 95.62 & 95.17 \\ \hline
M.T. & 76.55 & 80.18 & 78.15 & 81.04 & 51.49 & 58.46 & 95.26 & 94.54 \\
above + D. & 74.62 & 78.33 & 69.97 & 76.48 & 52.18 & 56.78 & 95.06 & 95.33 \\\hline
% above + D. & 74.55 & 78.12 & 52.17 & 59.26 & 95.53 & 94.77 \\\hline
M.T.+ P.A. & 77.51 & 80.64 & 78.41 & 81.42 & 58.09 & 58.93 & 94.88 & 94.91 \\
above + D. & 76.8 & 80.44 & 77.76 & 81.30 & 57.98 & 58.99 & 94.72 & 94.93 \\
\bottomrule
\end{NiceTabular}
}
\vspace{-3pt}
\caption{Ablation Study for Generative Annotator based on InternLM-7B in different settings.
Here, `S.T.' means single-task training, which only includes hallucination annotation task in training, while `M.T.' adopts multi-task training, which further encompasses several generative tasks.
``+ D'' indicates that testing the annotations with prompt disturbance \textit{i.e.}, the instructions used in testing are unseen in training. 
``P.A.'' indicates prompt augmentation is adopted in training.~\protect\footnotemark}
\label{tab:auto_result ann internlm2}
\vspace{-3pt}
\end{table}
\footnotetext{Due to the space limit, we put BERTScore in Tab~\ref{tab:ablation study internlm2 bert}.}

\noindent\textbf{Data Augmentation}
As shown in the first two rows of Tab.~\ref{tab:auto_result ann internlm2}, results are superior in the mix-task setting (introduced in \S~\ref{subsec: generative annotator}) compared to the single-task setting. This suggests that LLMs benefit from the multi-task shared representations and instruction-following ability.

% \noindent\textbf{Disturbance}
In addition, to evaluate the robustness of generative annotators, we introduce disturbance by altering the test instruction descriptions, ensuring they differ from the training instructions.
We compare the results obtained without and with prompt augmentation without and with disturbance in the last four rows of Tab.~\ref{tab:auto_result ann internlm2}. 
% Here, disturbance refers to alterations in the instructions in prompts for testing and train sets.
The model trained with prompt argumentation declines due to perturbations, less than that with augmentation (0.39\% vs. 6.37\% in accuracy).
It reveals models trained on diverse prompt formats increase robustness compared to their single prompt format-trained counterparts.

\begin{table}[!t]
\centering
\resizebox{1\linewidth}{!}{
\begin{NiceTabular}{lcccccccc}
\CodeBefore
  \cellcolor[HTML]{EFEFEF}{2-2,3-2,4-2,5-2,6-2,2-4,3-4,4-4,5-4,6-4,2-6,3-6,4-6,5-6,6-6,2-8,3-8,4-8,5-8,6-8}
\Body
\toprule
\multicolumn{1}{c}{\multirow{2}{*}{\textbf{Model}}} & \multicolumn{2}{c}{\textbf{F1 w/ Ref}}  & \multicolumn{2}{c}{\textbf{ACC w/ Ref}} & \multicolumn{2}{c}{\textbf{\textbf{F1 w/o Ref}}} & \multicolumn{2}{c}{\textbf{\textbf{ACC w/o Ref}}} \\ \cline{2-9}
\multicolumn{1}{c}{}   & \textbf{T}   & \textbf{Q}  & \textbf{T}   & \textbf{Q}   & \textbf{T}   & \textbf{Q}  & \textbf{T}   & \textbf{Q} \\ \hline
  G-7B & 75.93 & 77.24 & 77.89 & 78.12 & 52.86 & 55.84 & 57.34 & 58.69 \\
  G-20B & 79.82 & 81.18 & 80.21 & 81.81 & 58.06 & 59.95 & 59.51 & 61.2 \\  \hline
  D-7B & 66.20 & 68.53 & 69.15 & 70.86 & 57.24 & 59.84 & 60.15 & 61.32\\
  D-20B & 69.74 & 73.98 & 72.10 & 75.95 & 60.26 & 61.85 & 63.75 & 64.37 \\
  \bottomrule
\end{NiceTabular}
}
\vspace{-3pt}
\caption{Evaluation results for generative and discriminative annotators. Here, ``w/ Ref'' means providing reference documents when annotating, while ``w/o Ref'' means without reference documents.}
\label{tab: discriminative annotator result}
\vspace{-9pt}
\end{table}

\noindent\textbf{Reference}
We further examine the effectiveness of reference documents to the performance of the generative and discriminative annotators when judging the hallucination type. We test the annotators by compelling the model to rely solely on its parametric internal knowledge without any references.
% \ziwei{To initial the classification layer, the final linear layer of the discriminative annotator, we extract the weights for the label tokens from the trained annotator in \S~\ref{sec: annotator}.} 
Tab.~\ref{tab: discriminative annotator result} reveals that only relying on its parametric knowledge decreases the prediction \modified{F1 and accuracy}, indicating the importance of reference in annotating hallucinations.

\section{Related Work}
% 对第二段的展开
% 1. detector 未来可用于benchmark 2. mitigation
% 3页结束 1.5栏}
\noindent\textbf{Hallucination Benchmarks} can be broadly divided into two categories. %task-oriented and annotation-oriented 
% Two main types of Hallucination benchmarks currently exist: 
% (1) \textbf{Task-Oriented}: build challenging benchmarks in one/more NLP tasks and measure the hallucination level in the generated text. %~\cite{lin-etal-2022-truthfulqa, elaraby2023halo, Umapathi2023Med}.
% (2) \textbf{Annotation-Oriented}: train a hallucination detector/annotator with the benchmark and use the detector to score the hallucination degree. %~\cite{muhlgay2023generating,varshney2023stitch,Chern2023FacTool}. 
One type of benchmark mainly constructs challenging queries in one/multiple tasks and then evaluates the hallucination level in the responses~\citep{lin-etal-2022-truthfulqa, dziri2022faithdial,dziri2022origin,dziri2021evaluating,rohrbach2018object, li2024dawn}.
% TruthfulQA contains 817 challenging questions spanning 38 categories that some humans would answer falsely due to false beliefs or misconceptions.
% With the rapid progress of LLMs, early benchmarks, benchmarks tailored for LLMs have been introduced.
% This benchmark considers two types of hallucinations: imitative falsehoods and factual errors.
There are also domain-specific benchmarks curated recently, such as sports~\cite{elaraby2023halo} and medical~\cite{Umapathi2023Med} domains.
Besides these English benchmarks, a Chinese benchmark, HalluQA~\cite{cheng2023evaluating}, designs 450 adversarial questions spanning multiple domains.
While these benchmarks lean toward arising hallucinations, \methodname{} aims to provide an analytical framework for hallucination annotation.
% of hallucinations to facilitate further systematical mitigation.
% 不仅仅是trigger hallucination 提供了分析架构
% The limitations of these benchmarks in the data scale and comprehensiveness (in language and domains) impede the deeper analysis of the LLM hallucinations; thus, we construct \methodname{} that consists of a broad range of domains, including human, places, things, and events in both Chinese and English scenarios.

Another type of benchmarks can be used to train a hallucination detector/annotator and evaluate the hallucination level via the detector/annotator~\citep{liu2021token, dziri2022faithdial, gupta2022dialfact,laban2022summac,durmus2020feqa,wang2020asking, li2023halueval, varshney2023stitch, yang2023new, muhlgay2023generating}.
All these works classify the whole response of LLMs as either hallucinatory or not. Such a coarse-grained nature makes it difficult to conduct more detailed statistical analysis.
On the contrary, \methodname{} annotates hallucination for each sentence to different hallucination types with correction based on the retrieved reference documents.
Furthermore, \methodname{} collects natural responses from LLMs instead of artificially guiding LLMs to produce hallucinatory responses~\cite{li2023halueval, muhlgay2023generating}.
% include Chinese

% In light of these limitations including the lack of detailed analysis, unnatural responses, and limited language coverage, we propose \methodname{}. Our dataset contains analytical annotation of hallucinations in the naturally generated answers by LLMs, covering both Chinese and English.

% inhibit further development in hallucination detection research.
% is a collection of ChatGPT-generated and human-annotated hallucinated QA samples for evaluating LLMs' ability to recognize hallucinations. 
% FACTOR~\cite{muhlgay2023generating} defines a multi-choice task consisting of a prefix, one factually correct completion, and three non-factual completions generated according to different factual error types.
% \citet{varshney2023stitch} identifies the important concepts, calculates the model’s uncertainty on them, and then creates binary validation questions of the uncertain concepts.
% FactTool~\cite{Chern2023FacTool} leverages various tools, including Google Search, code interpreters, or even LLMs themselves, to gather evidence about the factuality of the generated content.
% \citet{yang2023new} propose a new benchmark for passage-level hallucination detection.

\paragraph{Hallucination Mitigation}
In the training stage, various techniques such as multi-task learning~\cite{weng2020towards, garg2019jointly}, model editing~\cite{daheim2023elastic, ji-etal-2023-rho}, and fine-grained RLHF~\cite{wu2023fine} are proposed to mitigate hallucination.
For inference time mitigation, different decoding strategies~\cite{rebuffel2022controlling, chuang2023dola, shi2023trusting,li2023inference} are attempted. 
There are also multi-agent methods~\cite{multiagentdebate2023} and variants of the Chain-of-Thought approach involving verification or reflection~\cite{dhuliawala2023chain,lei2023chain,ji2023towards,wang2023unleashing} proposed for LLMs.
The hallucination annotators trained on \methodname{} have the potential to be integrated into the training and inference pipeline by offering fine-grained hallucination information for further mitigation.

\section{Conclusion and Future Work}
Hallucinations in generative tasks present substantial obstacles to the reliability and creditability of LLMs but lack a comprehensive and fine-grained detecting strategy.
Thus, we present a bilingual dataset, \methodname{} for fine-grained hallucination annotation in GQA covering diverse topics, offering the opportunity to quantitatively analyze hallucination phenomena such as accumulation effect, and facilitating the development of state-of-the-art fine-grained hallucination annotators. 
Our generative hallucination annotators surpass all open-source LLMs and GPT-3.5 and obtain performance on par with GPT-4. 
Our generalization experiments indicate that improving the breadth of topics in the dataset is more important than extending questions under existing topics in the dataset. 

This paper paves the way for further scaling up the dataset of \methodname{} to conduct a systematic evaluation and analysis of LLM hallucinations, with the trained hallucination annotators. The hallucination annotators also have the potential to be used in the hallucination mitigation pipeline in both the training and inference stages.
% For future research, we will investigate underlying hallucination origins, especially those stemming from pre-training data. 
% Furthermore, we plan to broaden our exploration into the various stages of response generation in real-world applications. 
% For example, LLMs should have the capacity to refuse to answer adversarial questions, invoke tools for complex tasks, and seek external knowledge. % for unknown knowledge. 
% This will allow us to gain a more comprehensive understanding of the importance of each ability.

\section{Limitations}
This benchmark primarily incorporates the widely recognized and representative knowledge-intensive task, GQA. However, it does not encompass other tasks such as summarization and dialogue. 
During the dataset construction, we use GPT-3.5 with a reference document to construct a high-quality answer and an early version of InternLM-7B without reference to generate low-quality answers, respectively. Different models are used in that stage, we will further complete and analyze the other settings including GPT-3.5 without reference and InternLM-7B with reference.

In addition, our focus predominantly lies on the answer generation stage, without considering other stages such as the model's ability to recognize adversarial questions~\cite{Kumar2023Certifying,zhu2023promptbench}, red teaming~\cite{Ganguli2022Red}, acknowledge unknown knowledge~\cite{yin-etal-2023-large, rajpurkar2018know, Amayuelas2023Knowledge}, and retrieve accurate external knowledge once they realize their parametrical knowledge is not enough.

\section{Ethical Considerations}
We used publicly available reference documents for our benchmarks, effectively circumventing any possible harm toward individuals or groups.
The generated data by LLMs were carefully selected and processed by humans to secure privacy and confidentiality. No personal identification information was involved, and all data were made anonymous before any analysis was conducted. 
% We use ChatGPT to polish the writing and assist with language.

% \section*{Acknowledgements}

\bibliography{custom}

\appendix
\setcounter{table}{0} 
\setcounter{figure}{0}
\setcounter{equation}{0}
\renewcommand{\thetable}{A\arabic{table}}
\renewcommand\thefigure{A\arabic{figure}} 
\renewcommand\theequation{A\arabic{equation}}

\section{Dataset Construction}
\label{appendix sec: prompt}
\subsection{Topic Selection and Reference Retrieval} 
\begin{figure*}[t] 
\begin{AIbox}{}
{\bf English Prompt:} \\
{I will provide two entries along with introductions. Please determine if the two entries are synonymous, i.e., if the two entries refer to the same event, object, person, or location, etc. 

Entry 1: \{name1\}

Introduction 1: \{doc1\}

Entry 2: \{name2\}

Introduction 2: \{doc2\}

Are the two entries synonymous?}
\end{AIbox} 
\caption{Prompts for Reference Retrieval.}
\label{fig: reference retrieval prompt}
\end{figure*}

\modified{We use InternLM to assess whether a query and its candidate entries are synonymous via the prompt in Figure~\ref{fig: reference retrieval prompt}.}

\modified{If the sentence similarity between two questions exceeds the threshold, we consider them overly similar. The threshold is 300 for Chinese (via CoSENT) and 0.9 for English (via MiniLM), which are selected by case study.}

\subsection{Question Generation and Selection} 

First, we generate multiple questions based on the reference documents via prompts in Figure~\ref{fig: QG prompt}.

\begin{figure*}[t] 

\begin{AIbox}{}
{\bf English Prompt:} \\
{
I would like you to act as a question generator. I will provide references and you will generate 10 questions about "\{topic\}" based on the reference. The specific requirements are as follows:

1. the questions can be fully answered based only on the reference document, i.e. the answers to the questions are fully contained in the reference document. The questions should be objective and not too subjective or open-ended.

2. the 10 questions should be of as many different types as possible, e.g. what, when, where, why. Questions can be asked from different perspectives, e.g. descriptions, explanations, reasons, etc. Ensure that the questions are of different types and cover all aspects of the information.

3. 10 questions can cover different levels of knowledge, from general, basic knowledge to more specialized, complex subject knowledge or domain knowledge.

4. have only one question per item.

Reference: \{reference document\}

Please list the 10 questions directly based on the above reference without any explanation:
}\\\\
{\bf Chinese Prompt:} \\
{\begin{CJK}{UTF8}{gbsn} 我希望你充当一个问题生成器。我将提供参考资料，你将根据资料生成关于“\{topic\}”的10个问题。具体要求如下：

1. 只根据参考资料，完全可以回答问题，即问题的答案完全包含在参考资料中。问题要客观，不要太过主观和开放。

2. 10个问题尽量是不同类型的，比如：什么、何时、何地、为什么。问题可以从不同的角度出发，例如描述、解释、原因等。确保问题类型多样，覆盖资料的各个方面。

3. 10个问题可以涉及不同层次的知识，从常识性、基本性的知识，到更专业化、复杂化的学科知识或领域知识。

4. 每条只有一个问题。

参考资料： \{reference document\}

请根据以上参考资料，不做说明直接列出10个问题：
\end{CJK}}
\end{AIbox} 
\caption{Prompts for Question Generation.}
\label{fig: QG prompt}
\end{figure*}

We use GPT-3.5 to filter the open-ended subjective questions and make sure of their answerability via the prompts in Figure~\ref{fig: answerability prompt}.

\begin{figure*}[t] 

\begin{AIbox}{}
{\bf English Prompt:} \\
{I would like you to act as a question judge. Given several questions, determine if each question meets all of the following conditions: objective, about facts, has a definitive answer, and not open-ended.

\{questions\}

Please answer "yes" or "no" in label order, separated by line breaks and without any explanation.}\\
{\bf Chinese Prompt:} \\
{\begin{CJK}{UTF8}{gbsn}我希望你充当一个问题判断器。分别判断下列问题是否满足以下所有条件：客观的、关于事实的、有确切答案的、非开放的。

\{questions\}

请按标号顺序回答“是”或“否”，用换行符隔开，不加任何解释说明。
\end{CJK}} \\\\
{\bf English Prompt:} \\
{I would like you to act as a question answerability judge. I will provide a question and reference document, and you will judge whether the question is fully answerable based only on the reference document, i.e., whether the answer is included in the reference. 

Reference document: \{reference document\}

Question: \{question\}

Is it possible to answer the question at all, based only on the reference document? Please answer "yes" or "no" directly without any explanation.}\\
{\bf Chinese Prompt:} \\
{\begin{CJK}{UTF8}{gbsn}我希望你充当一个问题可回答性判断器。我将提供问题和参考资料，你将判断只根据参考文档，是否完全可以回答问题，即答案是否包含在参考资料中。

参考文档：\{reference document\}

问题：\{question\}

只根据参考文档，是否完全可以回答问题？请直接回答“是”或“否”，不加任何解释说明。
\end{CJK}}

\end{AIbox} 
\caption{Prompts for Question Answerability Judge.}
\label{fig: answerability prompt}
\end{figure*}

We use GPT-4 to select the final questions based on authenticity, answerability, difficulty, and variety via prompts in Figure~\ref{fig: best q prompt}.

\begin{figure*}[t] 

\begin{AIbox}{}
% {Prompts used in Question Selection}
{\bf English Prompt:} \\
{Good questions have the following characteristics:
1. high degree of truthfulness: the question contains no intentionally misleading, ambiguous or false information.
2. high answerability: remove questions that are too subjective, controversial, or predictive.
3. have a certain level of difficulty for the model.
4. increase the overall diversity (in terms of type, complexity, depth of knowledge, etc.), and remove questions that are similar to other questions.
Combine the above evaluation metrics and select the 3 best problems among these. Please respond directly to the question numbers, separated by commas, without any explanation.}\\\\
{\bf Chinese Prompt:} \\
{\begin{CJK}{UTF8}{gbsn} 好的问题具有以下特征：
1. 真实度高：问题中有没有故意误导、含糊不清或者虚假的信息。
2. 可回答性高：去掉过于主观、有争议、预测类的问题。
3. 对于模型有一定的难度。
4. 增加整体的多样性（类型、复杂度、知识深度等方面）,去除和其他问题相似的问题。
综合以上评价指标，在这些问题中选择3个最好的问题。请直接回复问题编号，用逗号隔开，不加任何解释说明。
\end{CJK}}
\end{AIbox} 
\caption{Prompts for Question Selection.}
\label{fig: best q prompt}
\end{figure*}

\subsection{Answering under Different Models and Scenarios} 
\label{appendix sec: answer prompt}
We generate answers with the document via prompts in Figure~\ref{fig: answer prompt}.

\begin{figure*}[t] 

\begin{AIbox}{}
{\bf English Prompt:} \\
{Reference document: \{reference document\}

Please answer the question based on the above reference: \{question\}}\\\\
{\bf Chinese Prompt:} \\
{\begin{CJK}{UTF8}{gbsn} 参考资料：\{reference document\}

请根据以上参考资料，回答问题：\{question\}
\end{CJK}}
\end{AIbox} 
\caption{Prompts for Answering.}
\label{fig: answer prompt}
\end{figure*}

\subsection{Fine-grained Hallucination Annotation} 
\label{appendix sec: annotation prompt}

We utilize GPT-4 to generate fine-grained hallucination annotation via prompts in Figure~\ref{fig: ann prompt0} to~\ref{fig: ann prompt4}.

\begin{figure*}[t] 
\begin{AIbox}{}
{\bf English Prompt:} \\
{I would like you to act as a hallucination annotator in an answer. I will provide a reference document and a question about "\{name\}" and you will judge whether the answer point contains hallucinations. The specific requirements are as follows:

1. If the point is supported by and consistent with the reference document, please write <Hallucination> None. And write the specific reference segment: <Reference> XXX. If there are multiple reference segments, please use "<SEP>" to separate them. Reference segments should be copied directly from the original text without modification.

2. If the point contradicts the reference document, please write: <Hallucination> Contradictory. And write the specific reference segment: <Reference> XXX. Also, write how to modify the answer: <Correction> "XXX" to "YYYY". If you need to delete XXX, write: <Correction> "XXX" to "".

3. If the point cannot be verified and there is no evidence in reference to support it, please write: <Hallucination> Unverifiable. And write the specific reference segment: <Reference> XXX. Also, write how to modify the answer: <Correction> "XXX" to "YYYY". If you need to delete XXX, write: <Correction> "XXX" to "".

4. If the point does not contain any factual information to be judged, please write: <No Fact>.

Question:  \{question\}

Reference: \{reference document\}

Point: \{answer sentence\}

Please annotate:}\\\\
{\bf Chinese Prompt:} \\
{\begin{CJK}{UTF8}{gbsn} 我希望你充当一个回答中的幻觉标注器。我将提供关于“\{name\}”的参考资料和问题，你将判断回答的要点是否含有幻觉。具体要求如下：

1. 如果要点与参考文档一致，请写：<幻觉>无。并注明参考片段：<参考>XXX。如果有多个参考片段，请用“<SEP>”分隔。参考片段应直接从原文复制，不需修改。

2. 如果要点与参考文档矛盾，请写：<幻觉>矛盾。并注明参考片段：<参考>XXX。同时说明如何修改回答：<改正>“XXX”改为“YYY”。如需删除内容XXX，请写：<改正>将“XXX”改为“”。

3. 如果要点无中生有，找不到证据支撑，无法验证，请写：<幻觉>无法验证。并注明参考片段：<参考>XXX。同时说明如何修改回答：<改正>“XXX”改为“YYY”。如需删除内容XXX，请写：<改正>将“XXX”改为“”。

4. 如果要点不包含待判断的事实信息，请写：<无事实>。

问题：\{question\}

参考文档：\{reference document\}

回答要点：\{answer sentence\}

请标注:\end{CJK}}
\end{AIbox} 
\caption{Prompts for Fine-grained Hallucination Annotation.}
\label{fig: ann prompt0}
\end{figure*}

\begin{figure*}[t] 
\begin{AIbox}{}
{\bf English Prompt:} \\
{I would like you to act as a hallucination annotator in an answer. I will provide a reference document and a question about "\{name\}" and you will judge whether the answer point contains hallucinations. The specific requirements are as follows:

1. If the point is supported by and consistent with the reference document, please write <Hallucination> None. And write the specific reference segment: <Reference> XXX. If there are multiple reference segments, please use "<SEP>" to separate them. Reference segments should be copied directly from the original text without modification.

2. If the point contradicts the reference document, please write: <Hallucination> Contradictory. And write the specific reference segment: <Reference> XXX. Also, write how to modify the answer: <Correction> "XXX" to "YYYY". If you need to delete XXX, write: <Correction> "XXX" to "".

3. If the point cannot be verified and there is no evidence in reference to support it, please write: <Hallucination> Unverifiable. And write the specific reference segment: <Reference> XXX. Also, write how to modify the answer: <Correction> "XXX" to "YYYY". If you need to delete XXX, write: <Correction> "XXX" to "".

4. If the point does not contain any factual information to be judged, please write: <No Fact>.

Reference: \{reference document\}

Question: \{question\}

Answer: \{answer sentence\}

Please annotate:}\\\\
{\bf Chinese Prompt:} \\
{\begin{CJK}{UTF8}{gbsn} 我希望你充当一个回答中的幻觉标注器。我将提供关于“\{name\}”的参考资料和问题，你将判断回答的要点是否含有幻觉。具体要求如下：

1. 如果要点与参考文档一致，请写：<幻觉>无。并注明参考片段：<参考>XXX。如果有多个参考片段，请用“<SEP>”分隔。参考片段应直接从原文复制，不需修改。

2. 如果要点与参考文档矛盾，请写：<幻觉>矛盾。并注明参考片段：<参考>XXX。同时说明如何修改回答：<改正>“XXX”改为“YYY”。如需删除内容XXX，请写：<改正>将“XXX”改为“”。

3. 如果要点无中生有，找不到证据支撑，无法验证，请写：<幻觉>无法验证。并注明参考片段：<参考>XXX。同时说明如何修改回答：<改正>“XXX”改为“YYY”。如需删除内容XXX，请写：<改正>将“XXX”改为“”。

4. 如果要点不包含待判断的事实信息，请写：<无事实>。

参考文档：\{reference document\}

问题：\{question\}

回答要点：\{answer sentence\}

请标注:\end{CJK}}
\end{AIbox} 
\caption{Prompts for Fine-grained Hallucination Annotation.}
\label{fig: ann prompt1}
\end{figure*}

\begin{figure*}[t] 
\begin{AIbox}{}
{\bf English Prompt:} \\
{I would like you to act as a hallucination annotator in an answer. I will provide a reference document and a question about "{name}" and you will judge whether each point of the answer contains hallucinations. The specific requirements are as follows:

1. If the point does not contain any factual information to be judged, please write: <No Fact>. And end the annotation.

2. If the point contains factual information, please find the specific reference segment and write: <Reference> XXX. If there are multiple reference segments, please use "<SEP>" to separate them. Reference segments should be copied directly from the original text without modification.

3. If the point is supported by and consistent with the reference document, please write: <Hallucination> None.

4. If the point contradicts the reference document, please write: <Hallucination> Contradictory. Also, write how to modify the answer: <Correction> "XXX" to "YYYY". If you need to delete XXX, write: <Correction> "XXX" to "".

5. If the point cannot be verified and there is no evidence in reference to support it, please write: <Hallucination> Unverifiable. Also, write how to modify the answer: <Correction> "XXX" to "YYYY". If you need to delete XXX, write: <Correction> "XXX" to "".

Question:  \{question\}

Reference: \{reference document\}

Please annotate the answer:\{answer sentence\}}\\\\
{\bf Chinese Prompt:} \\
{\begin{CJK}{UTF8}{gbsn} 我希望你充当一个回答中的幻觉标注器。我将提供关于“{name}”的参考资料和问题，你将判断回答的每个要点是否含有幻觉。具体要求如下：

1. 如果要点不包含待判断的事实信息，请写：<无事实>，并结束标注。

2. 如果要点包含事实信息，请找相关的参考片段，请写：<参考>XXX。如果有多个参考片段，请用“<SEP>”分隔。参考片段应直接从原文复制，不需修改。

3. 如果要点与参考文档一致，请写：<幻觉>无。

4. 如果要点与参考文档矛盾，请写：<幻觉>矛盾。同时说明如何修改回答：<改正>“XXX”改为“YYY”。如需删除内容XXX，请写：<改正>将“XXX”改为“”。

5. 如果要点无中生有，找不到证据支撑，无法验证，请写：<幻觉>无法验证。同时说明如何修改回答：<改正>“XXX”改为“YYY”。如需删除内容XXX，请写：<改正>将“XXX”改为“”。

问题：\{question\}

参考文档：\{reference document\}

请标注要点：\{answer sentence\}\end{CJK}}
\end{AIbox} 
\caption{Prompts for Fine-grained Hallucination Annotation.}
\label{fig: ann prompt2}
\end{figure*}

\begin{figure*}[t] 
\begin{AIbox}{}
{\bf English Prompt:} \\
{Imagine you are a detective who specializes in identifying hallucinations. I will provide you with reference documents and questions about "{name}" and you will need to evaluate each point of information in the responses for the presence of hallucinations.  Please follow the steps below:

- If the information point does not contain a fact that can be judged, mark: <No Fact> and end the annotation.

- If the information point contains a fact, list the corresponding reference: <Reference> XXX. If there is more than one, separate them with "<SEP>". Please ensure that the reference information is copied directly from the original text and does not need to be altered.

- If the information point is consistent with the reference, please mark: <Hallucination> None.

- If the information point contradicts the reference, please mark it as <Hallucination> Contradictory and include a correction: <Correction> "XXX" to "YYYY". When something needs to be eliminated, write: <Correction> "XXX" to "".

- If the information point cannot find relevant evidence, or cannot be verified, please mark: <Hallucination> Unverifiable, and include a correction: <Correction>"XXX" to "YYYY". When you need to eliminate something, please write: <Correction> "XXX" to "".

Question:  \{question\}

Reference: \{reference document\}

Please annotate the information point: \{answer sentence\}}\\\\
{\bf Chinese Prompt:} \\
{\begin{CJK}{UTF8}{gbsn} 想象你是一个专门鉴别幻觉的侦查员。我将向你提供关于“{name}”的参考文档和问题，你需要评估回答中的每个信息点是否存在幻觉。请按以下步骤进行：

- 如信息点不包含可判断的事实，请标明：<无事实>，并结束评估。

- 如信息点包含事实，请列出相应的参考信息点：<参考>XXX。若有多个，请以“<SEP>”分隔。请确保参考信息直接复制自原文，无需更改。

- 如信息点与参考内容一致，请标注：<幻觉>无。

- 如信息点与参考内容相矛盾，请标注：<幻觉>矛盾，并附上改正方法：<改正>“XXX”改为“YYY”。需要剔除某内容时，请写：<改正>将“XXX”改为“”。

- 如信息点无法找到相关证据，或无法验证，请标注：<幻觉>无法验证，并附上改正方法：<改正>“XXX”改为“YYY”。需要剔除某内容时，请写：<改正>将“XXX”改为“”。

问题：\{question\}

参考文档：\{reference document\}

请标注信息点：\{answer sentence\}\end{CJK}}
\end{AIbox} 
\caption{Prompts for Fine-grained Hallucination Annotation.}
\label{fig: ann prompt3}
\end{figure*}

\begin{figure*}[t] 
\begin{AIbox}{}
{\bf English Prompt:} \\
{You are now a hallucination detection system. I will provide you with a reference document and a question on the topic "{name}". Your task is to analyze the responses to the question and determine whether or not there is a hallucination for each point. The steps of the assessment are as follows:

- If it does not contain factual information that needs to be judged, write: <No Fact> and stop the assessment.

- If facts are included, identify the relevant reference clip. Write: <Reference> XXX. Separate multiple references with "<SEP>". Please copy the reference fragment directly from the original without modification.

- If the points are identical to the reference, write: <Hallucination> None.

- If the main points are contradictory to the reference document, write: <Hallucination> Contradictory. Include a suggestion for revision: <Correction> "XXX" to "YYY". If a section needs to be deleted, write: <Correction> "XXX" to "".

- If no evidence can be found to support a point, or if it cannot be verified, write: <Hallucination> Unverifiable, with a suggested change: <Correction> "XXX" to "YYYY". If a section needs to be deleted, write: <Correction> "XXX" to "".

Question:  \{question\}

Reference: \{reference document\}

Please analyze the point:\{answer sentence\}}\\\\
{\bf Chinese Prompt:} \\
{\begin{CJK}{UTF8}{gbsn} 你现在是一个幻觉检测系统。我会为你提供关于主题“{name}”的一篇参考文档和一个问题。你的任务是分析问题的回答，判断每个要点是否存在幻觉。评估步骤如下：

- 如果没有包含需要判断的事实信息，请写：<无事实>，并停止评估。

- 如果包含事实，找出相关参考片段。请写：<参考>XXX。多个参考片段请用"<SEP>"分隔。参考片段请直接从原文复制，不要修改。

- 如果要点与参考完全一致，请写：<幻觉>无。

- 如果要点与参考文档存在矛盾，写：<幻觉>矛盾。并附上修改建议：<改正>“XXX”改为“YYY”。如果需要删除某部分，写：<改正>将“XXX”改为“”。

- 如果无法找到证据支持要点，或无法验证，写：<幻觉>无法验证，并附上修改建议：<改正>“XXX”改为“YYY”。如果需要删除某部分，写：<改正>将“XXX”改为“”。

问题：\{question\}

参考文档：\{reference document\}

请分析要点：\{answer sentence\}\end{CJK}}
\end{AIbox} 
\caption{Prompts for Fine-grained Hallucination Annotation.}
\label{fig: ann prompt4}
\end{figure*}

\section{Case Study}
\label{appendix sec: Case}
Table~\ref{tab: negative case1},~\ref{tab: negative case2}, and~\ref{tab: negative case3} show the examples where the GPT-4 generated annotation is inconsistent with human annotation.

\begin{table*}[!ht]
\centering
 \resizebox{1\linewidth}{!}{
\begin{tabular}{ll}
\toprule
\textbf{Topic} & \multicolumn{1}{c}{Fear of crime}  \\ \hline
% 4a3659e0-3871-46d9-8d4b-0eec6aa929d0
\textbf{Reference}   & \begin{tabularx}{1\textwidth}{X}\textbf{Reference 0:} 
The fear of crime refers to the fear of being a victim of crime as opposed to the actual probability of being a victim of crime. The fear of crime, along with fear of the streets and the fear of youth, is said to have been in Western culture for "time immemorial". While fear of crime can be differentiated into public feelings, thoughts and behaviors about the personal risk of criminal victimization, distinctions can also be made between the tendency to see situations as fearful, the actual experience while in those situations, and broader expressions about the cultural and social significance of crime and symbols of crime in people's neighborhoods and in their daily, symbolic lives...

\# Interpersonal communication and the mass media

Hearing about events and knowing others who have been victimised are thought to raise perceptions of the risk of victimisation. This has been described as a 'crime multiplier', or processes operating in the residential environment that would 'spread' the impacts of criminal events... {\color[HTML]{3166FF}A subject's criminal risk perception is exaggerated by peer-communication on crime and only moderated by the own experience. Public perceptions of the risk of crime are no doubt also strongly shaped by mass media coverage.}
\end{tabularx}   \\\hline
\textbf{Question}   & \begin{tabularx}{1\textwidth}{X}  What are the three different components of risk perception that may interact in relation to the likelihood of falling victim to crime? \end{tabularx} \Tstrut\Bstrut  \\  \hline
\textbf{\begin{tabular}[c]{@{}l@{}}Selected\\ Ans Sent \end{tabular}} &  \begin{tabularx}{1\textwidth}{X}
The three different components of risk perception that may interact in relation to the likelihood of falling victim to crime are: 

1. Perceived Risk: This refers to an individual's subjective assessment of the likelihood of being a victim of crime.
{\color[HTML]{3166FF}It is influenced by factors such as personal experience, media coverage, and social norms.}

2. Actual Risk: This refers to the objective probability of being a victim of crime, based on statistical data and crime patterns.

3. Perceived Benefits and Costs: This refers to an individual's assessment of the benefits and costs of taking precautions to reduce their risk of being a victim of crime.

It is influenced by factors such as the perceived effectiveness of different precautions, the perceived costs of taking precautions, and the perceived benefits of not taking precautions.
\end{tabularx} \\ \hline
\textbf{\begin{tabular}[c]{@{}l@{}}GPT-4\\ Annotation\end{tabular}} &  \begin{tabularx}{1\textwidth}{X} \textbf{<No Fact>} \end{tabularx}  \\ \hline

\textbf{\begin{tabular}[c]{@{}l@{}}Human\\ Annotation\end{tabular}} \Tstrut\Bstrut &  \begin{tabularx}{1\textwidth}{X}  \textbf{<Reference>} Reference 0: A subject's criminal risk perception is exaggerated by peer-communication on crime and only moderated by the own experience. Public perceptions of the risk of crime are no doubt also strongly shaped by mass media coverage.

\textbf{<Halluciantion>} None  \end{tabularx}  \\ 
\bottomrule   
\end{tabular}
}
  \caption{An example for inconsistent GPT-4 and human annotation.}
  \label{tab: negative case1}
\end{table*}

\begin{table*}[!ht]
\centering
 \resizebox{1\linewidth}{!}{
\begin{tabular}{ll}
\toprule
\textbf{Topic} & \multicolumn{1}{c}{Neo-Freudianism}  \\ \hline
% 8c55deed-8a85-4333-b741-80d81e323e85
\textbf{Reference}   & \begin{tabularx}{1\textwidth}{X}\textbf{Reference 0:} 
Neo-Freudianism is a psychoanalytic approach derived from the influence of Sigmund Freud but extending his theories towards typically social or cultural aspects of psychoanalysis over the biological...Karen Horney theorized that to deal with basic anxiety, the individual has three options: Moving toward others: Accept the situation and become dependent on others.  This strategy may entail an exaggerated desire for approval or affection. {\color[HTML]{3166FF}Moving against others: Resist the situation and become aggressive. This strategy may involve an exaggerated need for power, exploitation of others, recognition, or achievement.} Moving away from others: Withdraw from others and become isolated. This strategy may involve an exaggerated need for self-sufficiency, privacy, or independence... \end{tabularx}   \\\hline
\textbf{Question}   & \begin{tabularx}{1\textwidth}{X} What are the three options that Karen Horney theorized individuals have to deal with basic anxiety?\end{tabularx} \Tstrut\Bstrut  \\  \hline
\textbf{\begin{tabular}[c]{@{}l@{}}Selected\\ Ans Sent \end{tabular}} &  \begin{tabularx}{1\textwidth}{X}
Karen Horney theorized that individuals have three options to deal with basic anxiety: 

1. Moving towards people: This option involves seeking out social connections and relationships in order to feel safe and secure.
Individuals who choose this option may become overly dependent on others and struggle with feelings of inadequacy.

2. Moving against people: This option involves competing with others and striving for power and control in order to feel safe and secure.
{\color[HTML]{3166FF}Individuals who choose this option may become overly aggressive and struggle with feelings of insecurity.}

3. Moving away from people: This option involves withdrawing from social connections and relationships in order to feel safe and secure.
Individuals who choose this option may become overly detached and struggle with feelings of loneliness and isolation.
\end{tabularx} \\ \hline
\textbf{\begin{tabular}[c]{@{}l@{}}GPT-4\\ Annotation\end{tabular}} &  \begin{tabularx}{1\textwidth}{X}
\textbf{<Halluciantion>} Unverifiable  

\textbf{<Correction>} "Individuals who choose this option may become overly aggressive and struggle with feelings of insecurity." to "Karen Horney theorized that to deal with basic anxiety, the individual has three options: Moving toward others (becoming dependent on others), moving against others (becoming aggressive), or moving away from others (becoming isolated)."
\end{tabularx} \Tstrut\Bstrut \\ \hline

\textbf{\begin{tabular}[c]{@{}l@{}}Human\\ Annotation\end{tabular}} &  \begin{tabularx}{1\textwidth}{X}  \textbf{<Reference>} Reference 0: Moving against others: Resist the situation and become aggressive. This strategy may involve an exaggerated need for power, exploitation of others, recognition, or achievement.

\textbf{<Halluciantion>} None  \end{tabularx} \Tstrut\Bstrut \\ 
\bottomrule   
\end{tabular}
}
  \caption{An example for inconsistent GPT-4 and human annotation.}
  \label{tab: negative case2}
\end{table*}

\begin{table*}[!ht]
\centering
 \resizebox{1\linewidth}{!}{
\begin{tabular}{ll}
\toprule
\textbf{Topic} & \multicolumn{1}{c}{Psychosis}  \\ \hline
% 6bb25edb-754d-4d99-b2a6-bfe88e649a17
\textbf{Reference}   & \begin{tabularx}{1\textwidth}{X}\textbf{Reference 0:} 
Psychosis, any of several major mental illnesses that can cause delusions, hallucinations, serious defects in judgment and other cognitive processes, and the inability to evaluate reality objectively... {\color[HTML]{3166FF} Other medical conditions associated with psychoses include thyroid disease, vitamin deficiencies, liver disease, epilepsy, Parkinson disease, and encephalitis.} Disorders of sleep or prolonged sensory deprivation can also cause hallucinations and delusions. [1,6]Many drugs can alter brain function and cause symptoms of psychoses.  \end{tabularx}   \\\hline
\textbf{Question}   & \begin{tabularx}{1\textwidth}{X} How are hallucinations related to abnormal brain activity? \end{tabularx} \Tstrut\Bstrut  \\  \hline
\textbf{\begin{tabular}[c]{@{}l@{}}Selected\\ Ans Sent \end{tabular}} &  \begin{tabularx}{1\textwidth}{X}
Hallucinations are a symptom of abnormal brain activity.
Hallucinations can be caused by a variety of factors, including mental illness, substance abuse, and neurological disorders.
When the brain is not functioning properly, it can produce hallucinations as a result of abnormal activity in the visual, auditory, or other sensory systems.
For example, in schizophrenia, the brain's dopamine system is overactive, leading to hallucinations and delusions.
Similarly, in Parkinson's disease, abnormal activity in the basal ganglia can cause visual hallucinations.
{\color[HTML]{3166FF}Other conditions, such as epilepsy and migraines, can also cause hallucinations as a result of abnormal brain activity.}
Overall, hallucinations are a symptom of abnormal brain activity and can be caused by a variety of underlying conditions.
\end{tabularx} \\ \hline
\textbf{\begin{tabular}[c]{@{}l@{}}GPT-4\\ Annotation\end{tabular}} &  \begin{tabularx}{1\textwidth}{X}
 \textbf{<Reference>} Reference 0: Other medical conditions associated with psychoses include thyroid disease, vitamin deficiencies, liver disease, epilepsy, Parkinson disease, and encephalitis. Disorders of sleep or prolonged sensory deprivation can also cause hallucinations and delusions.
 
\textbf{<Halluciantion>} None  
\end{tabularx} \Tstrut\Bstrut \\ \hline

\textbf{\begin{tabular}[c]{@{}l@{}}Human\\ Annotation\end{tabular}} &  \begin{tabularx}{1\textwidth}{X}  \textbf{<Reference>} Reference 0:  Other medical conditions associated with psychoses include thyroid disease, vitamin deficiencies, liver disease, epilepsy, Parkinson disease, and encephalitis. 

\textbf{<Halluciantion>} Unverifiable

\textbf{<Correction>} "and migraines" to "".
\end{tabularx} \Tstrut\Bstrut \\ 
\bottomrule   
\end{tabular}
}
  \caption{An example for inconsistent GPT-4 and human annotation.}
  \label{tab: negative case3}
\end{table*}

\section{Implementation Details}
\label{appendix sec: implementation}
\subsection{Generative Annotator}
\label{appendix sec: generative ann implementation}
% \ziwei{Among them, the stream2 and additional pad are significantly better than the default stage concatenation method
% Data ratio
% Does it count as input loss or not? The effect is good
% And model bases, etc}

The maximum sequence length is set to 16k. This setting is also held constant in baselines. 
We load the pre-trained InternLM2-7B model and train it with the following settings and hyper-parameters: the epoch is 1, the batch size is 2, the learning rate is 4e-5, and the AdamW optimizer is with a linear scheduler. 
We generate responses using sampling implemented via the LMDeploy library\footnote{\url{https://github.com/InternLM/lmdeploy}}. 
Our model is trained on 8 NVIDIA A800 GPUs. It takes approximately 1 hour to train.

\subsection{Discriminative Annotator}
\label{appendix sec: discriminative ann implementation}
We use InternLM2-7B and 20B as the base model for training. 
We train the discriminative annotator on our benchmark with the following settings and hyper-parameters: the epoch is 2, the batch size is 8, the learning rate is 1e-5, the AdamW optimizer is with a linear scheduler, and the maximum sequence length is 16k.
Our model is trained on 8 NVIDIA A800 GPUs.

\section{Results and Analysis}
\label{appendix sec: results}

\begin{table}[!t]
\centering
\resizebox{1\linewidth}{!}{
\begin{tabular}{lcccccc}
\toprule
\multicolumn{1}{c}{\textbf{Topic}} & \textbf{F1} $\uparrow$ & \textbf{ACC} $\uparrow$ & \textbf{R} $\uparrow$ & \textbf{BERT} $\uparrow$ & \textbf{Pre4} $\uparrow$ \\ \hline \Tstrut
Person & 75.8 & 76.58 & 53.69 & 87.69 & 52.79 \\
Event & 70.48 & 73.33 & 52.26 & 80.70 & 42.94 \\
Location & 83.34 & 83.81 & 75.39 & 92.77 & 67.86 \\
Thing & 79.36 & 80.23 & 58.19 & 87.34 & 29.53 \\
\bottomrule
\end{tabular}
}
\vspace{-6pt}
\caption{Topic-specific automatic evaluation results for generative hallucination annotators \methodname{}-7B, where `R', `BERT', and `Pre4' refer to `RougeL', `BERTScore', and `4-gram Precision', respectively.}
\label{tab: topic-specific performance}
\vspace{-6pt}
\end{table}

\modified{Topic-specific automatic evaluation results for generative hallucination annotators are shown in Tab.~\ref{tab: topic-specific performance}. The trained \methodname{}-7B performs best on location topics while struggling with event topics.}

\begin{table}[!t]
\centering
% \resizebox{1\linewidth}{!}{
\begin{NiceTabular}{lcc}
\CodeBefore
  \cellcolor[HTML]{EFEFEF}{2-2,3-2,4-2,5-2}
\Body
\toprule
\multicolumn{1}{c}{\multirow{2}{*}{\textbf{Setting}}} & \multicolumn{2}{c}{\textbf{BERT}$\uparrow$} \\ \cline{2-3} 
\multicolumn{1}{c}{}   & \textbf{T}   & \textbf{Q} \\ \hline 
\begin{tabular}[c]{@{}l@{}}
G-7B\end{tabular} &  87.29 & 87.27  \\
G-20B & 87.96 & 88.93 \\ 
\bottomrule
\end{NiceTabular}
% }
% \quad\quad
\vspace{-6pt}
\caption{Evaluation results for generative annotators, noted by `G'. `T' represents the unseen-topic test set, while `Q' represents the unseen-question test set.}
\label{tab:gen_ann}
\end{table}

\begin{table}[!t]
\centering
% \resizebox{1\linewidth}{!}{
\begin{NiceTabular}{lcc}
\CodeBefore
  \cellcolor[HTML]{EFEFEF}{2-2,3-2,4-2,5-2,6-2,7-2}
\Body
\toprule
\multicolumn{1}{c}{\multirow{2}{*}{\textbf{Setting}}} & \multicolumn{2}{c}{\textbf{BERT}$\uparrow$} \\ \cline{2-3} 
\multicolumn{1}{c}{}   & \textbf{T}   & \textbf{Q} \\ \hline 
\begin{tabular}[c]{@{}l@{}}
S.T.\end{tabular} & 87.29 & 87.27 \\ \hline
M.T. & 85.94 & 87.55 \\
above + D. & 86.05 & 86.71 \\\hline
% above + D. & 85.99 & 87.6 \\\hline
M.T.+ P.A. & 86.95 & 87.6 \\
above + D. & 86.89 & 87.58 \\
\bottomrule
\end{NiceTabular}
% }
\vspace{-6pt}
\caption{Ablation Study for Generative Annotator based on InternLM-7B in different settings.
Here, `S.T.' means single-task training, which only includes hallucination annotation task in training, while `M.T.' adopts multi-task training, which further encompasses several generative tasks.
``+ D'' indicates that testing the annotations with prompt disturbance \textit{i.e.}, the instructions used in testing are unseen in training. 
``P.A.'' indicates prompt augmentation is adopted in training.}
\label{tab:ablation study internlm2 bert}
\end{table}

Figure~\ref{fig:cm_ann} shows the confusion matrices of hallucination type for annotators in different sizes.
Figure~\ref{fig:cm_scoring_7} and~\ref{fig:cm_scoring_20} show the confusion matrices for discriminative annotators under different scenarios in different sizes.

\begin{figure}[!ht]
 \centering
 \includegraphics[width=1\linewidth]{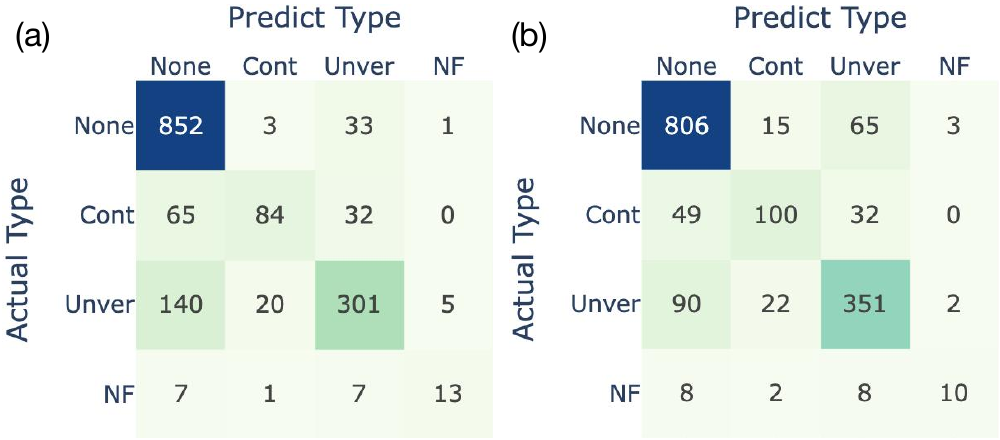}
  \caption{Hallucination Type Confusion Matrices for Generative Annotators. (a) InternLM2-7B-based annotator (b) InternLM2-20B-based annotator}
  \label{fig:cm_ann}
  \vspace{-1.1 em}
\end{figure}

\begin{figure}[!ht]
 \centering
 \includegraphics[width=1\linewidth]{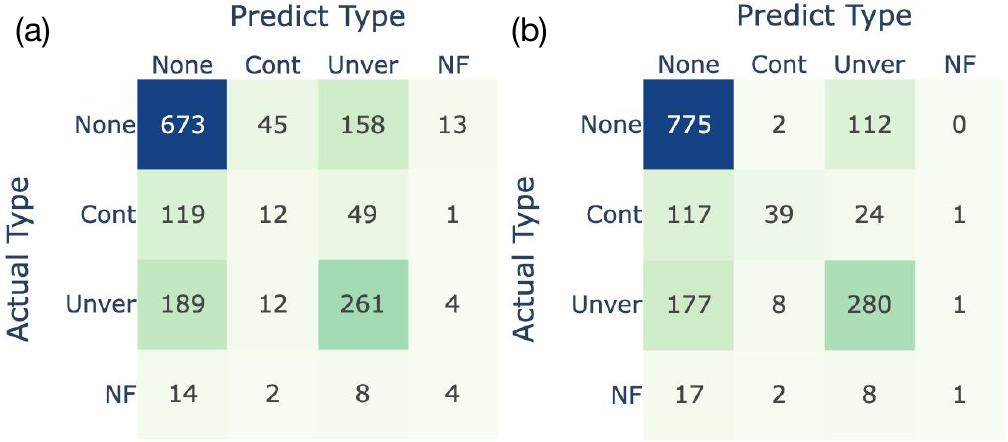}
  \caption{Hallucination Type Confusion Matrices for Discriminative Annotators based on InternLM2-7B. (a) without reference (b) with reference}
  \label{fig:cm_scoring_7}
  \vspace{-1.1 em}
\end{figure}

\begin{figure}[!ht]
 \centering
 \includegraphics[width=1\linewidth]{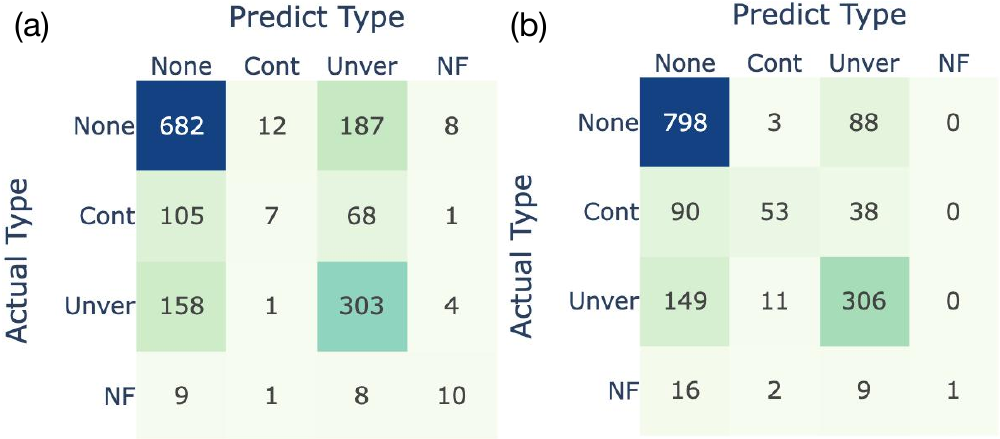}
  \caption{Hallucination Type Confusion Matrices for discriminative annotators based on InternLM2-20B. (a) without reference (b) with reference}
  \label{fig:cm_scoring_20}
  \vspace{-1.1 em}
\end{figure}

\subsection{Generalization on other LLMs}
\label{appendix sec: generalization on other LLMs}

\begin{table}[!t]
\centering
% \resizebox{1\linewidth}{!}{
\begin{tabular}{lcc}
\toprule
\textbf{Model} & \textbf{F1}$\uparrow$ & \textbf{ACC}$\uparrow$  \\\hline
Qwen-7B & 73.80 & 76.76  \\
Baichuan2-7B & 75.97 & 77.55 \\ 
\bottomrule
\end{tabular}
% }
% \quad\quad
\vspace{-6pt}
\caption{Evaluation results for generative annotators.}
\label{tab:gen_ann_other}
\vspace{-15pt}
\end{table}

\modified{To assess generalizability, we sample ~100 sentence-level annotations for answers generated by other models (Qwen-7B, Baichuan2-7B). We manually check the quality of \methodname{}-7B annotator as in Table~\ref{tab:gen_ann_other}. The accuracy for other models is similar to that of GPT3.5 and InternLM. It proves that our annotator is still relatively stable on other models.}

\modified{We find that for the same query, generated answers from different models are around the topic and they are not far apart. Thus, in the context of factual QA, the divergence is not substantial and the answers are relatively in domain. Please find some examples in Tab.~\ref{tab: other LLMs}.}

\begin{table*}[!ht]
\centering
\begin{tabularx}{\textwidth}{lX}
\toprule
\textbf{Question} & \multicolumn{1}{l}{How did Victor Hugo gain wider fame in 1831?}  \\ \hline
\textbf{InternLM2-7B} & Victor Hugo gained wider fame in 1831 with his historical novel Notre-Dame de Paris (The Hunchback of Notre-Dame), an evocation of life in medieval Paris during the reign of Louis XI... \\ \hline
\textbf{Qwen-7B} & Victor Hugo gained wider fame in 1831 with his historical novel Notre-Dame de Paris, which is an evocation of life in medieval Paris during the reign of Louis XI... \\\hline
\textbf{Baichuan2-7B} & Victor Hugo gained wider fame in 1831 when he published his novel Les Mis\'erables. This novel was published in 1831 and was well received by critics and the public... \\ 
\bottomrule 

\end{tabularx}
\caption{Generated answers from different models for the same query.}
  \label{tab: other LLMs}
\end{table*}

\section{Human Annotation}
\label{appendix sec: Annotation}
The annotation platform is developed internally by the laboratory. Human annotators, comprising well-educated undergraduates. Their salary is 300 yuan per day which is adequate given the participants' demographic. An ethics review board approved the data collection protocol.

Human annotation involves two stages: (1) screening topics and references; and (2) fine-grained hallucination annotation. We provide comprehensive instructions for each task, including task descriptions, precautions, estimated time, three examples, and three negative cases, to facilitate understanding.

\modified{We also employ a double annotation process during human annotation:
(1) Annotators fix the GPT4 pre-annotations. 
(2) Experienced annotators (selected by platform) review the annotations and give feedback.
Multiple rounds of (1) and (2) are performed until the platform deems the annotation acceptable.
(3) NLP experts check the annotation quality to finally decide whether to accept it. The pass rate is ~85\% and unqualified samples are redone until accepted.}

% The UI for human annotation is shown in Figue~\ref{fig:UI}.
% \begin{figure}[!ht]
%  \centering
%  \includegraphics[width=1\linewidth]{figures/UI.pdf}
%   \caption{The example of fine-grained hallucination annotation with the UI.}
%   \label{fig:UI}
%   \vspace{-1em}
% \end{figure}

\end{document}